\documentclass[10pt,twocolumn,letterpaper]{article}

\usepackage{cvpr}              

\usepackage{graphicx}
\usepackage{amsmath}
\usepackage{amssymb}
\usepackage{booktabs}

\usepackage{arydshln}
\usepackage{multirow}
\usepackage[accsupp]{axessibility}  

\usepackage[pagebackref,breaklinks,colorlinks]{hyperref}

\usepackage[capitalize]{cleveref}
\crefname{section}{Sec.}{Secs.}
\Crefname{section}{Section}{Sections}
\Crefname{table}{Table}{Tables}
\crefname{table}{Tab.}{Tabs.}

\def\cvprPaperID{5298} 
\def\confName{CVPR}
\def\confYear{2023}

\begin{document}

\newcommand\correspondingauthor{\thanks{Corresponding author.}}

\title{Adaptive Assignment for Geometry Aware Local Feature Matching}
\author{Dihe Huang$^{1}$\footnotemark[1] \quad
        Ying Chen$^{2}$\footnotemark[1] \quad
        Yong Liu$^{2}$ \quad
        Jianlin Liu$^{2}$ \quad
        Shang Xu$^{2}$ \\ 
        Wenlong Wu$^{2}$ \quad
        Yikang Ding$^{1}$ \quad
        Fan Tang$^{4}$\footnotemark[2] \quad
        Chengjie Wang$^{2,3}$\footnotemark[2] \\
        $^{1}$Tsinghua Univeristy \quad $^{2}$Tencent YouTu Lab \quad $^{3}$ Shanghai Jiao Tong University\\
        $^{4}$ Institute of Computing Technology, Chinese Academy of Sciences\\
        {\tt\small \{hdh20, dyk20\}@mails.tsinghua.edu.cn \quad tfan.108@gmail.com} \\
        {\tt\small \{mumuychen, choasliu, jenningsliu, shangxu, wenlongwu, jasoncjwang\}@tencent.com}}

\maketitle
\renewcommand{\thefootnote}{\fnsymbol{footnote}}
\footnotetext[1]{These authors contributed equally.}
\footnotetext[2]{Corresponding author.}

\begin{abstract}
The detector-free feature matching approaches are currently attracting great attention thanks to their excellent performance. However, these methods still struggle at large-scale and viewpoint variations, due to the geometric inconsistency resulting from the application of the mutual nearest neighbour criterion (\ie, one-to-one assignment) in patch-level matching.
Accordingly, we introduce AdaMatcher, which first accomplishes the feature correlation and co-visible area estimation through an elaborate feature interaction module, then performs adaptive assignment on patch-level matching while estimating the scales between images, and finally refines the co-visible matches through scale alignment and sub-pixel regression module.
Extensive experiments show that AdaMatcher outperforms solid baselines and achieves state-of-the-art results on many downstream tasks. Additionally, the adaptive assignment and sub-pixel refinement module can be used as a refinement network for other matching methods, such as SuperGlue, to boost their performance further. The code will be publicly available at https://github.com/AbyssGaze/AdaMatcher.
\end{abstract}
    
\section{Introduction}
\label{sec:intro}

\begin{figure}[ht]
    \centering
    \includegraphics[width=0.46\textwidth]{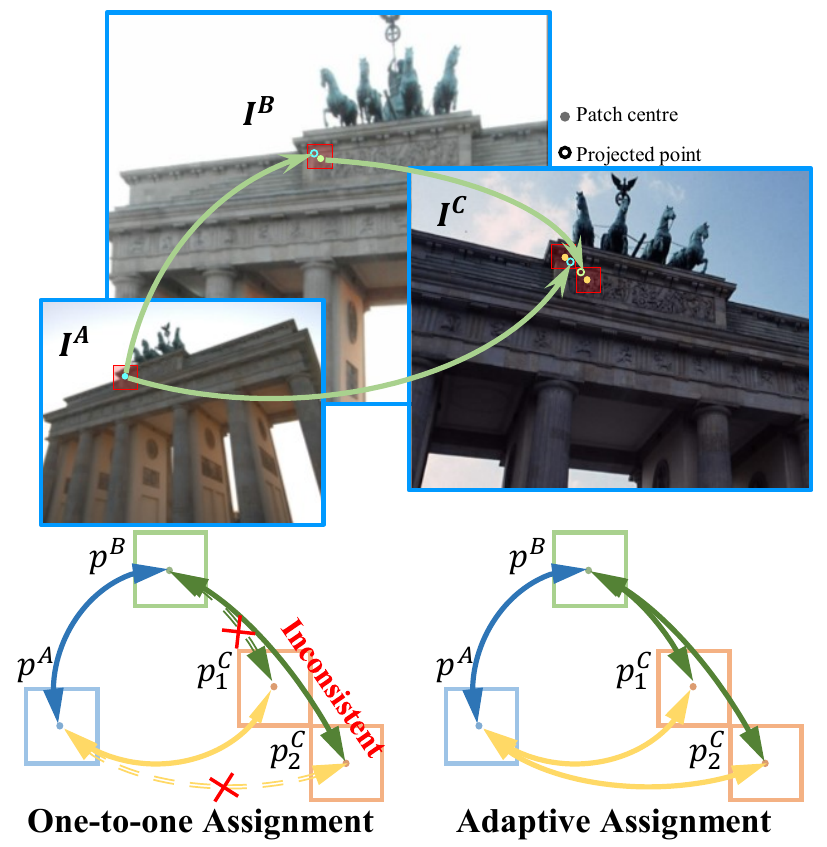}
    \caption{\textbf{An illustration of one-to-one assignment and adaptive assignment.} Under viewpoint changes or scale variations, one-to-one assignment leads to geometric inconsistency in patch-level feature matching, while adaptive assignment does not. 
    For example, with one-to-one assignment, patch pair $(p^A, p^C_2)$ is treated as a negative example, even though both $p^C_1$ and $p^C_2$ are projected into $p^A$ of $I^A$. Such a matching rule is inconsistent with two-view and multi-view projective geometry. 
    }
    \label{introduction}
\end{figure}

Establishing accurate correspondences for local features between image pairs is an essential basis for a broad range of computer vision tasks, including visual localization, structure from motion (SfM), simultaneous localization and mapping (SLAM), etc. However, achieving reliable and accurate feature matching is still challenging due to various factors such as scale changes, viewpoint diversification, illumination variations, repetitive patterns, and poor texture.

Existing image matching pipelines are mainly divided into two types: detector-based and detector-free. The former is to build matches on detected and described sparse keypoints~\cite{Lowe2004sift, Rublee2011, Detone2018superpoint, revaud2019r2d2, Luo2020aslfeat, tyszkiewicz2020disk}. However, as the detector-based matching pipeline relies on the reliability of keypoint detectors and features description, it tends to perform poorly under large viewpoint changes or scale variations. For the latter, the detector-free matching pipeline can take full advantage of the rich context to establish correspondence between images end-to-end \cite{Rocco2018, Rocco2020, Li2020, truong2021pdcnet, JiamingSunHujunBao2021loftr, tang2021quadtree, chen2022aspanformer}, without independent keypoint detection and feature description steps. 
To achieve efficiency and accurate matching, the SOTA detector-free matching pipelines~\cite{Li2020, jiang2021cotr, JiamingSunHujunBao2021loftr, tang2021quadtree, chen2022aspanformer} use a coarse-to-fine structure, in which the patch-level matches are first obtained using the mutual nearest neighbor criterion, and then are refined to a sub-pixel level.

Although these methods have improved considerably in performance, they still perform unsatisfactorily in extreme cases (\eg, large viewpoint changes and scale changes). 
This is due to the fact that applying the mutual nearest neighbor criterion (ie, one-to-one correspondence) in patch-level matching leads to geometric inconsistencies and difficulties in obtaining sufficient high-quality matches under large-scale or viewpoint variations. As shown in Fig.\ref{introduction}, where $I^A, I^B, I^C$ are from the same scene, $p^C_1$ and $p^C_2$ of $I^C$ are both projected into $p_A$ of $I^A$. 
However, when the mutual nearest neighbour criterion is applied in the training process, the patch pair $(p^A, p^C_1)$ is treated as a positive sample, while the patch pair $(p^A, p^C_2)$ is treated as a negative sample. The incorrect assignment leads to two-view geometric inconsistency. Deeply, from a multi-view perspective, $(p^A, p^B)$ and $(p^B, p^C_2)$ are positive samples while $(p^A, p^C_2)$ is a negative sample, which leads to multi-view geometric inconsistency between multiple image pairs.
For inference, when there are large viewpoint changes or scale variations, one-to-one matching is difficult to obtain enough inliers to ensure accurate camera pose estimation. Furthermore, when applied to multi-view-based downstream tasks (\eg, SfM and 3D reconstruction), one-to-one patch-level correspondences do not guarantee the consistency of multi-view matching, which is likely to make the mapping fail or the bundle adjustment difficult to converge.

Inspired by the above consideration, we present AdaMatcher, a geometry aware local feature matching approach, targeting at mitigating potential geometry mismatch between image pairs without scale-alignment preprocessing or viewpoint warping. Different from dual-softmax or optimal transport in \cite{JiamingSunHujunBao2021loftr, sarlin2020superglue} which guarantees one-to-one correspondence, we allow adaptive assignment (including many-to-one and one-to-one) at patch-level matching during training and inference. 
When the scale or viewpoint changes significantly, the adaptive assignment can guarantee matching accuracy.
The smooth scale transition from many-to-one matches between image pairs can be adopted to resolve scale inconsistencies. Furthermore, the structure of our delicately designed feature interaction module couples co-visible feature decoding with cross-feature interaction, allowing the probability map of the co-visible region to be obtained later by a simple module to filter out matches outside co-visible areas.
To summarize, we aim to provide several critical insights of matching local features across scales and viewpoints:
\begin{itemize}
\item We propose a detector-free matching approach AdaMatcher that allows a patch-level adaptive assignment followed by a sub-pixel refinement to guarantee the establishment of geometry aware feature correspondences. 
\item We introduce a novel feature interaction structure, which couples the co-visible feature decoding and cross-feature interaction. The probability map of the co-visible area can be obtained later by an additional module.

\item Extensive experiments and analysis demonstrate that AdaMatcher outperforms various strong baselines and achieves SOTA results for many downstream vision tasks.
\end{itemize}

\section{Related Work}

\begin{figure*}[ht]
    \centering
    \includegraphics[width=0.95\textwidth]{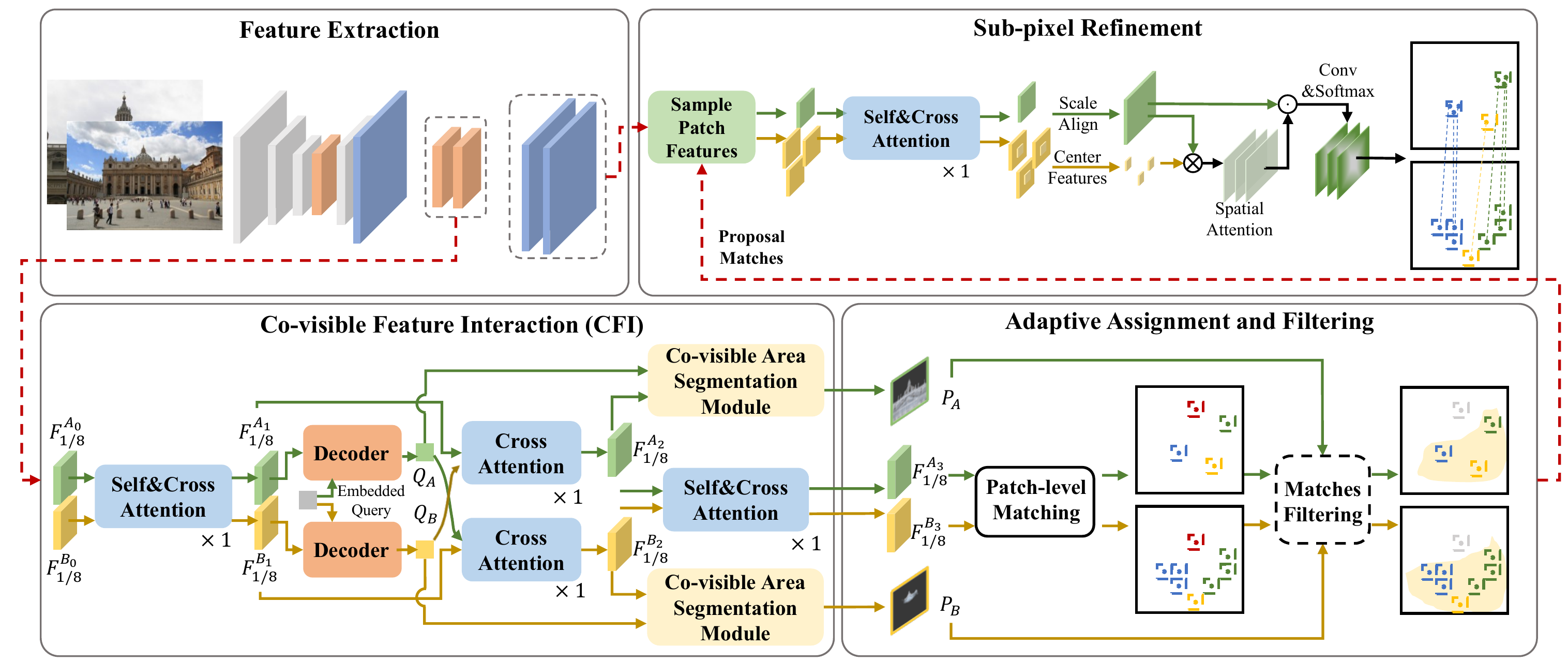}
    \caption{\textbf{Architecture of AdaMatcher.} A local feature CNN extracts two feature maps with size of 1/2 and 1/8 of the input image dimension. Afterwards, 1/8 size features of the two images are correlated by our feature interaction module, followed by an extra module to estimate the co-visible area shared between two images. Then the interacted patch features and co-visible probability maps are fed into an adaptive assignment and filtering module, yielding sufficient patch-level matches. Finally, proposal-matched features are sampled in the 1/2 size feature maps with a scale alignment to mitigate scale mismatch, followed by sub-pixel refinement.}
    \label{network}
\end{figure*}

\subsection{Scale- or Viewpoint-invariant Local Feature}
To tackle geometry deformation induced by scale and viewpoint variation across images, tremendous efforts~\cite{Lowe2004sift, Rublee2011, barroso2019key, revaud2019r2d2, barroso2020hdd, Luo2020aslfeat, liu2021densernet, Detone2018superpoint, YChen2022overlap, barroso2022scalenet} have been made within local feature matching pipelines. Hand-crafted local features such as SIFT \cite{Lowe2004sift} or ORB \cite{Rublee2011} adopt scale-space theory \cite{Lindeberg1998} to alleviate potential large-scale variations. However, descriptors extracted locally from low-level image textures possess poor discrimination ability. Recently, many works have been devoted to a learning-based approach to tackle local feature matching under scale variations or viewpoint changes. Methods directly performing convolution upon the multi-scale pyramid such as KeyNet~\cite{barroso2019key}, R2D2~\cite{revaud2019r2d2} and HDDNet~\cite{barroso2020hdd}, or implicitly applying multi-scale detection such as ASLfeat~\cite{Luo2020aslfeat} and DenseNet~\cite{liu2021densernet} are intended to mimic conventional scale space theory. 
However, the multi-scale pyramid brings the side effect of ambiguity since correspondence needs to be established among multiple scale levels. 
There are also works~\cite{Detone2018superpoint, pautrat2020online} aiming at invariance to different scales through an elaborate learning process, however, which would render them less discriminative~\cite{wiles2021co}. Some works~\cite{berton2021viewpoint, zaffar2021vpr} in geo-localization aim to achieve viewpoint invariance. GeoWarp~\cite{berton2021viewpoint} directly warps pairwise images to a closer geometrical space to eliminate viewpoint inconsistencies and then computes their similarities using dense local features for image retrieval tasks. In addition, OETR~\cite{YChen2022overlap} estimates overlap areas as a preprocessing module in the existing detector-based matching pipeline to constrain keypoint detection in the co-visible areas and eliminate scale and viewpoint inconsistencies. However, its need to scale up the whole image increases the time and computational consumption of the later feature extraction and matching steps. 
In contrast to the above approaches, our proposed method is inspired by many-to-one matching caused by viewpoint and scale variations. The adaptive assignment is introduced to handle feature matching under view and scale variations, and the relative scales between two images can be estimated from the results of patch-level adaptive matching, which can be used for feature scale alignment in the subsequent refinement step. 
In addition, inspired by the overlap estimation of OETR~\cite{YChen2022overlap}, we couple the co-visible feature decoding into feature interaction to make the network more focused on the co-visible regions, thus alleviating the performance degradation in extreme cases.

\subsection{Detector-free Image Matching}
Recent works~\cite{Rocco2018, Li2020, JiamingSunHujunBao2021loftr, truong2021pdcnet, tang2021quadtree, chen2022aspanformer} have shown us that end-to-end dense feature matching without keypoint detection can be more robust than detector-based matching methods in many scenarios.
NCNet~\cite{Rocco2018} and its follow-ups \cite{Rocco2020,Li2020} propose a 4D matching cost volume to enumerate all possible correspondences and obtain dense matches end-to-end. Although all the potential matches are considered in the 4D matching tensor, the receptive field of 4D convolution is still limited to each match’s neighborhood area. Benefiting from the global receptive field and long-range dependencies from Transformers, LoFTR\cite{JiamingSunHujunBao2021loftr} and its variants~\cite{tang2021quadtree, chen2022aspanformer} extend neighborhood consensus to the whole image, setting the SOTA performance for dense feature matching approaches. However, \cite{Rocco2018, Rocco2020, Li2020, truong2021pdcnet, JiamingSunHujunBao2021loftr, tang2021quadtree, chen2022aspanformer} do not handle the case of significant viewpoint and scale changes well because they follow one-to-one matching. In comparison, the use of adaptive assignment can make the dense feature matching methods more robust in extreme cases.

\section{Methods}
This section describes our proposed matching framework, named AdaMatcher, as shown in Fig.\ref{network}.
Given a pair of images $I^A$ and $I^B$, we first feed them into a CNN backbone to obtain coarse features and fine features. Then, the coarse features are passed through our CFI module (Sec.~\ref{CFI}) to accomplish feature Interaction and co-visible area estimation. After that, adaptive assignment~(Sec.\ref{m2o}) is applied to get the patch-level matches and calculate the relative scales between image pairs, while the previously estimated co-visible regions are used to filter the matches. Finally, the patch-level matches are scale-aligned and refined to sub-pixel precision (Sec.~\ref{refine}) according to the estimated scale.

\subsection{Co-visible Feature Interaction} \label{CFI}
Our ultimate goal is to bring existing detector-free matching methods to be more robust under scale or viewpoint changes. We find that overlap estimation~\cite{YChen2022overlap} helps to improve the matching performance in extreme cases. However, introducing a full network for co-visible region estimation would be computationally and time-consuming. Instead, we couple feature interactions with co-visible feature decoding so that co-visible features can be used to guide global feature interactions while reducing computation. On the one hand, co-visibility guidance can suppress features in non-co-visible regions, facilitating the subsequent matching step. On the other hand, a simple additional module can be used to obtain co-visible regions to filter mismatches.


\noindent{\bf Feature Interaction with Co-visible Feature Decoding.} 
As shown in the bottom left part of Fig.\ref{network}, we first use one set of self- and cross-attention layer as the feature encoder to acquire information within and across images. The output features are denoted as $F^{A_1}_{1/8}$ and $F^{B_1}_{1/8}$ respectively. 
For co-visible feature initialization, we adopt one query $Q{\in}\mathbb{R}^{1\times{d}}$ to embed co-visible context, where $d$ is the channel dimension, and perform one cross-attention layer to decode the locations of the co-visible region. $Q^{A}$ and $Q^{B}$ denote the features decoded from $F^{A_1}_{1/8}$ and $F^{B_1}_{1/8}$ respectively, i.e., $Q^{i}=\mathbf{transfomer}(q=Q, k=v=F^{i_1}_{1/8}), i\in\{A,B\}$. 
After that, co-visible features can be used to guide global feature interactions through a cross transformer, of which the outputs are denoted as $F^{A_2}_{1/8}$ and $F^{B_2}_{1/8}$ respectively.
Finally, to make the local features more distinguishable, we use another set of self- \& cross-attention layers to construct a complete graph for feature correlation. The proposed feature interaction structure can be directly applied to LoFTR~\cite{JiamingSunHujunBao2021loftr}, QuadTree~\cite{tang2021quadtree} and ASpanFormer~\cite{chen2022aspanformer}, \ie, the corresponding variants can be obtained according to different attention mechanisms.

\begin{figure}[ht]
    \centering
    \includegraphics[width=0.47\textwidth]{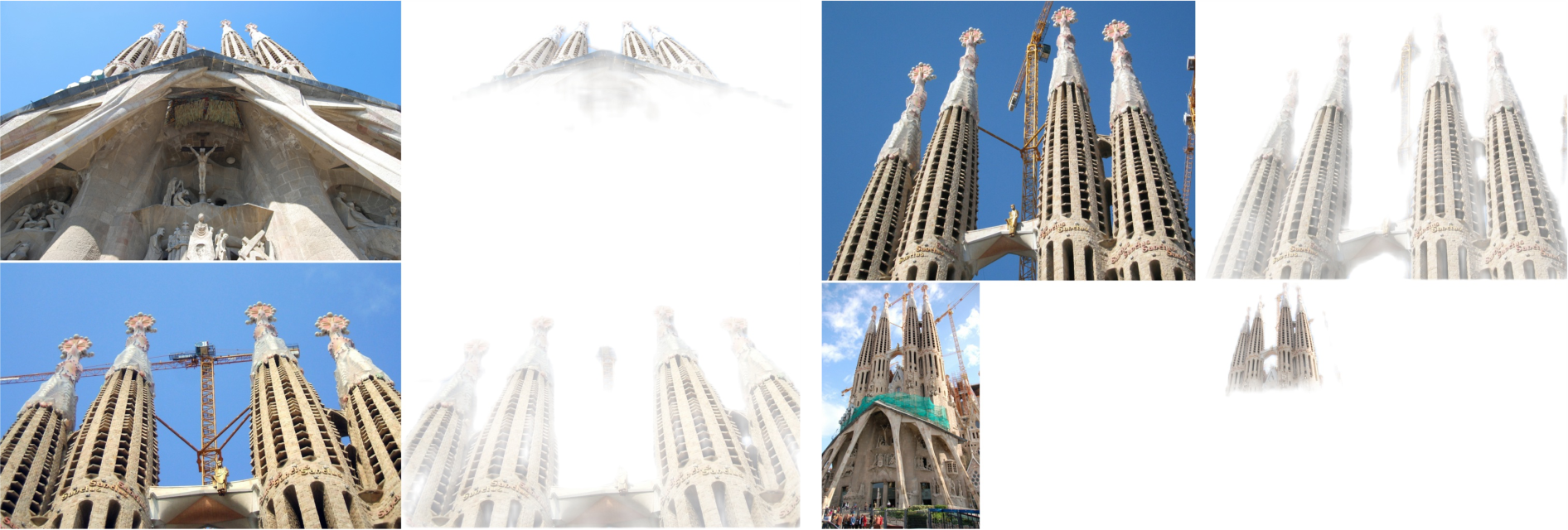}
    \caption{\textbf{Co-visible area segmentation visualizations.} Visualized with two different scenes, the first column is the origin image pair, and the second is the co-visible area.}
    \label{mask}
\end{figure}

\noindent{\bf Co-visible area segmentation.}
After co-visible feature decoding, a simple additional module can be used to obtain the co-viewing regions.
Here we consider co-visible area segmentation as predicting a logit probability map, whose values at each pixel represent the probability of being in the co-visible region, as shown in Fig.\ref{mask}. In detail, we project the decoded features $Q^{A, B}$ to construct a weight map using matrix multiplication and a sigmoid function. The weight map is used to enhance the co-visible context in feature maps $F^{A_2}_{1/8},F^{B_2}_{1/8}$. Then a convolution operation with the kernel size of $3\times3$ and a sigmoid function are applied, which is detailed in Eq.~\eqref{eq2}:
\begin{equation}
\begin{aligned}
& \mathbf{{weight}_{i}}=\mathbf{Sigmoid}({(F^{i_2}_{1/8})^T}Q^i),\\
& {P}_{i}=\mathbf{Sigmoid}(\mathbf{Conv}(\mathbf{{weight}_{i}}\odot{F^{i_2}_{1/8}}+F^{i_2}_{1/8})), \label{eq2}
\end{aligned}
\end{equation}
where $i  \in \{A,B\}$, $\odot$ denotes element-wise multiplication and ${P}_{A},{P}_{B}$ denote the co-visible probability map of image $A$ and image $B$. After obtaining ${P}_{A}$ and ${P}_{B}$, a confidence threshold can be applied to retain the co-visible areas in image $A$ and image $B$.

\subsection{Adaptive Assignment}\label{m2o}
\begin{figure*}[ht]
    \centering
    \includegraphics[width=0.7\textwidth]{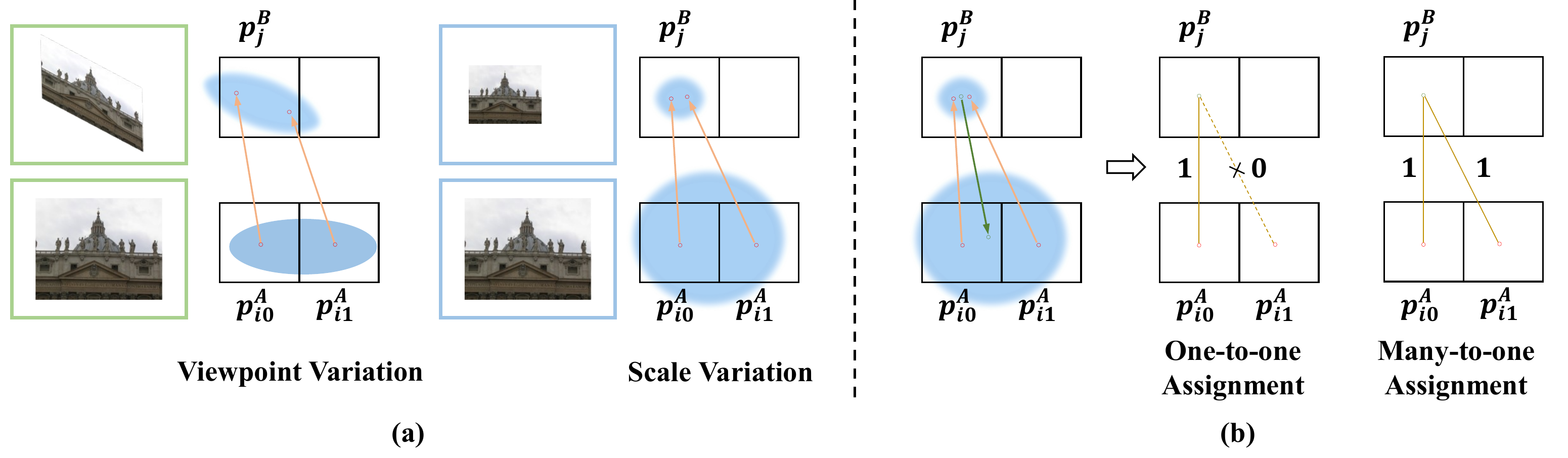}
    \caption{\textbf{Comparison of one-to-one and many-to-one assignment.} (a) shows patch-level many-to-one matching due to viewpoint and scale changes; (b) shows the difference between many-to-one and one-to-one assignment: one-to-one only keeps a single match while both $p_{i0}^A$ and $p_{i1}^A$ correspond to $p_j^B$, and many-to-one assignment keeps common area matches to disambiguate positive and negative samples, to resolve geometry deformation.}
    \label{fig4}
\end{figure*}
In this section, we will elaborate on one of the main contributions of our work: adaptive assignment when matching between features across images. 
As shown in Fig.\ref{fig4}(a), when scale varies or viewpoint changes, centers of several patches within one image would be projected into only one patch of the other image, named many-to-one correspondences. For the ground truth patch-level labels obtained by one-to-one assignment, only the correspondences that satisfy the mutual nearest neighbors constraint are taken as positive samples, while the others as negative samples. Such ambiguous label assignment is detrimental to supervised training. As shown in Fig.\ref{fig4}(b), while a set of patch centers (or pixels) of image $A$: $\{P^A_i|p^A_{ik}, k=1,2,...,N\}$ are all projected into a patch (or pixel) of image $B$: $p^B_j$ using ground-truth camera poses and depth maps, features corresponding to $\{P^A_i\}$ are similar to the feature corresponding to $p^B_j$. Following mutual nearest neighbors constraint, $({p}^A_{im},{p}^B_j)$ would be assigned as positive sample, while $\{({p}^A_{ik}, {p}^B_j)|k=1,2,...,N, k!=m \}$ are assigned as negative, where $m=\mathop{\arg\min}_{k}\|\mathrm{D}(\mathrm{W}(p^A_{ik}),  p^B_{j})\|$, $\mathrm{D}(\cdot)$ is the projected distance between matching candidates and $\mathrm{W}(\cdot)$ demonstrates the projection function. Such one-to-one assignment criterion will turn good correspondences into negative samples, which is inconsistent with the multi-view geometry theory. Instead, adaptive assignment will make correspondences $\{({p}^A_{ik}, {p}^B_j)|k=1,2,...,n\}$ being positive samples since their appearances are similar and they also conform to geometric constraint. When applied to multi-view tasks (\eg SfM), the one-to-one assignment cannot guarantee the multi-view geometric consistency. 
Instead, when adaptive many-to-one/one-to-many/one-to-one assignments are allowed at patch-level matching, feature inconsistency under large-scale or viewpoint changes will be mitigated.

\noindent\textbf{Matching matrix formulation.}
Given features $F^{A_3}_{1/8}$ and $F^{B_3}_{1/8}$ output from CFI module, we calculate their similarity matrix $\mathcal{S}$:
\begin{equation}
\begin{aligned}
\mathcal{S}(i,j)=\frac{1}{r}\cdot\left\langle F^{A_3}_{1/8}(i), F^{B_3}_{1/8}(j)\right \rangle,
\end{aligned}
\end{equation}
where $i$ and $j$ are index of feature patches in $I^{A}$ and $I^{B}$ respectively, and $\left \langle \cdot , \cdot \right \rangle $ denotes inner product. 
Adaptive assignment is a one-way operation that consists of many-to-one and one-to-one assignment, i.e., we assign "many" patches on images with a large co-visible area to "one" patch on images with a small co-visible area. When little scale or viewpoint variation exists, many-to-one assignment is adaptive to become one-to-one. Hence, we apply softmax operation to similarity matrix $\mathcal{S}(i,j)$ on two dimensions separately, followed by selecting similar matches with a threshold $\theta_{m}$:
\begin{equation}
\begin{aligned}
& \mathcal{P}_k = \mathbf{softmax}(\mathcal{S}(i, \cdot))_j, \\ 
& \mathcal{M}_k = \{(\widetilde{i},\widetilde{j})|\mathcal{P}_k(\widetilde{i},\widetilde{j}) > \theta_{m}\}, \\
\end{aligned}
\end{equation}
where $k \in {0, 1}$, $\mathcal{P}_0$ and $\mathcal{P}_1$ are the matching probability matrix obtained by softmax operation along the first dimension and the zeroth dimension, $\mathcal{M}_0$ and $\mathcal{M}_1$ are the corresponding patch-level match proposals. Then we select the matching probability matrix $\mathcal{P}$ and the proposal matches $\mathcal{M}$: $\mathcal{P} = \mathcal{P}_{index}, \mathcal{M} = \mathbf{Filtering}(\mathcal{M}_{index}, {P}_A, {P}_B, \theta_{co-visible})$,
where $\mathbf{Filtering}$ is to filter out matches outside predicted co-visible areas, $\theta_{co-visible}$ is used to select patches in the co-visible probability maps belong to the co-visible region, and $index = \mathop{\arg\max}_k \{s_k|k=0,1\}$. $s_k$ is the scale between images, calculated by 
\begin{equation}
\begin{aligned}
&s_k = \frac{\mathbf{len}(\mathcal{M}_k)}{\mathbf{len}(\mathbf{unique}(\mathcal{M}_k[:,1-k]))}, k=0,1.\\
\end{aligned}
\label{eq5}
\end{equation}

\subsection{Sub-pixel Refinement Module}\label{refine}
By adaptive assignment we obtain patch-level match proposals through scales and viewpoints. Then we refine these match proposals to more accurate sub-pixel level matches, by a scale-alignment and an expectation regression module.


\noindent\textbf{Scale-alignment.} 
Suppose scale for $p^A_i$ is larger than $p^B_j$, match proposals are $\{{p}^A_{i\in \Omega}, {p}^B_j\}$, and $\Omega$ is the collection of assigned patches. Patch features can be then sampled in $F^A_{1/2}$ and $F^B_{1/2}$, followed by one self/cross attention layer to communicate feature messages from assigned patches. The scale ratio is $s=max(s_0/s_1, s_1/s_0)$, where $s_0$ and $s_1$ could be calculated from Eq.~\eqref{eq5}. Smaller scale image features are upsampled by this $s$ to compensate for scale mismatch between images. 

\noindent\textbf{Sub-Pixel level regression.} 
To locate accurate sub-pixel level matches, we generate a heatmap representing the matching probability for each pixel. 
Firstly, we correlate center features of ${F}^A_{i\in \Omega}$ and the scale-alignment features ${F'}^B_j$ to calculate $n$ spatial attention maps. Then we perform a dot-product operation on the attention maps and ${F'}^B_j$ to balance the relevance for each features. Finally, a simple convolution and softmax is employed to predict probability distribution, from which the final position $i'$ with sub-pixel accuracy on $I_{A}$ is obtained by taking expectation over distribution. By this scale-aware adaption to match position refinement, we achieve more accurate sub-pixel matches.

\subsection{Supervision}
\noindent\textbf{Co-visible Area Segmentation Loss.}
For the co-visible area segmentation, we treat it as a per-pixel binary classification task. The loss $\mathcal{L}_{co-visible}$ can be calculated by Focal Loss \cite{Lin2017fl} (abbr. as $FL$ hereafter):
\begin{equation}
\mathcal{L}_{co-visible}=FL(P_A,\widehat{P}_A)+FL(P_B,\widehat{P}_B),
\end{equation}
where $\widehat{P}_A$ and $\widehat{P}_B$ denote the ground-truth co-visible areas of image $A$ and image $B$, respectively, which are calculated based on depth and camera poses.

\noindent\textbf{Proposal Matching Loss.} For the loss function of the adaptive matching probability matrix $\mathcal{P}$, we use the same Focal Loss \cite{Lin2017fl} as in LoFTR~\cite{JiamingSunHujunBao2021loftr}: 
\begin{equation}
\mathcal{L}_{M} = FL(\mathcal{P}, \widetilde{\mathcal{P}}),
\end{equation}
where $\widetilde{\mathcal{P}}$ is the ground-truth labels of the adaptive matching probability matrix calculated from the camera poses and depth maps, and $\mathcal{P}$ is the predicted matching probability matrix. For the $\alpha$ and $\gamma$ parameters in Focal Loss \cite{Lin2017fl}, we use the default values, which are set to 0.25 and 2, respectively. 
For patch-level label generation, we project the patch centroids of the two images onto each other using depth and camera poses and then characterize all points projected into the same patch as positive samples with that patch.

\noindent\textbf{Refinement Loss.}
Inspired by LoFTR\cite{JiamingSunHujunBao2021loftr}, we use the same loss $\mathcal{L}_{refine}=\frac{1}{|\mathcal{M}_{gt}^{k}|}\sum_{i, j'} \frac{1}{\sigma(i)^2} \left \| j'-j'_{gt} \right \|$ function for the final predicted matches, 
where $k$ is the $index$ calculated above. $M_{gt}^{k}$ is the ground-truth matches calculated from ground-truth depths and camera poses and $\sigma ^2(\cdot)$ is the variance of the corresponding heatmap \cite{JiamingSunHujunBao2021loftr}. 

Our final loss is balanced as:
\begin{equation}
\mathcal{L} = 0.5*\mathcal{L}_{co-visible} + 1.0*\mathcal{L}_{M} + 1.0*\mathcal{L}_{refine}.
\end{equation}

\subsection{Implementations}\label{Implementations}
We train AdaMatcher on the MegaDepth datasets following \cite{JiamingSunHujunBao2021loftr}, without any data augmentation. We apply AdaMatcher to LoFTR~\cite{JiamingSunHujunBao2021loftr} and its variants (QuadTree Attention~\cite{tang2021quadtree} and ASpanFormer~\cite{chen2022aspanformer}), named AdaMatcher-LoFTR, AdaMatcher-Quad and AdaMatcher-ASpan, respectively. The only difference between these three variants is the type of attention mechanism in the feature interaction module, which are linear attention~\cite{lineartransformer}, quadtree attention~\cite{tang2021quadtree} and attention span~\cite{chen2022aspanformer}. All networks are trained using AdamW optimizer with initial learning rate of $8 \times 10^{-3}$ and batch size of 8. It converges after 2 days of training on 8 V100 GPUs. The image feature extractor is a standard ResNet-FPN \cite{He2016,Lin2017} architecture, which is identical to LoFTR \cite{JiamingSunHujunBao2021loftr}. $\theta_m$ is set to 0.5,  $\theta_{co-visible}$ is set to 0.2 and the patch window size $w$ for refinement is set to 5. The number of channels of the $F_{1/8}$ and $F_{1/2}$ is 256 and 128, respectively. To save GPU memory usage during training, we sample 30 percent of matches (max to 2500) from the match proposals for supervision in sub-pixel refinement module. More details are provided in Supplementary Material A.

\section{Experiments}


\subsection{Homography Estimation}
\noindent\textbf{Dataset.} HPatches \cite{Balntas2017} is the most widely used image matching evaluation dataset. There are 116 scenes with 57 sequences of large illumination variations and 59 sequences under significant viewpoint changes to evaluate our method under different circumstances. All images are resized to their longer dimensions equal to 1024, and we limit the maximum amount of matches to 1K for all methods.

\noindent\textbf{Metrics.} Following \cite{Detone2018superpoint, sarlin2020superglue, zhou2021patch2pix} we use corner correctness to describe the performance of estimated homography. Four corners in the first reference image are wrapped to the other image by estimated homography. Then percentange of correct estimated homographies whose average error of the four corners is less than $1/3/5$ pixels demonstrates the matching \emph{Accuracy}. We use OpenCV RANSAC as the robust estimator following \cite{zhou2021patch2pix}. 

\begin{table}[]
\setlength\tabcolsep{3pt}
\centering

\resizebox{.46\textwidth}{!}{
\begin{tabular}{clll}
\toprule
\multirow{2}{*}{\textbf{Category}} &
  \multicolumn{1}{l}{\multirow{2}{*}{\textbf{Method}}} &
  \multicolumn{1}{c}{\textbf{Overall}} &
  \multicolumn{1}{c}{\textbf{Viewpoint}} \\ \cmidrule{3-4} 
                               & \multicolumn{1}{c}{} & \multicolumn{2}{c}{Accuracy($\epsilon$ \textless{}1/3/5px)}                     \\ \midrule
                               
\multirow{5}{*}{\textbf{Detector-}} 
& KeyNet\cite{barroso2019key}+HardNet\cite{mishchuk2017working} & \multicolumn{1}{c|}{0.30\ /0.61\ /0.75} &  \multicolumn{1}{c}{0.14\ /0.46\ /0.64\ }\\
\multirow{5}{*}{\textbf{based}} 
& SIFT\cite{Lowe2004sift}+HardNet\cite{mishchuk2017working} & \multicolumn{1}{c|}{0.33\ /0.59\ /0.74} &  \multicolumn{1}{c}{0.20\ /0.40\ /0.60\ }\\
& SP\cite{Detone2018superpoint} & \multicolumn{1}{c|}{0.31\ /0.66\ /0.78} &  \multicolumn{1}{c}{0.18\ /0.51\ /0.64\ } \\
& R2D2(MS)\cite{revaud2019r2d2} & \multicolumn{1}{c|}{0.29\ /0.60\ /0.72\ } & \multicolumn{1}{c}{0.18\ /0.43\ /0.58\ }  \\
& SP\cite{Detone2018superpoint}+CAPS\cite{Wang2020caps} & \multicolumn{1}{c|}{0.27\ /0.66\ /0.71\ }  & \multicolumn{1}{c}{0.15\ /0.53\ /0.65\ } \\
& Patch2Pix\cite{zhou2021patch2pix} & \multicolumn{1}{c|}{0.34\ /0.68\ /0.79\ }   & \multicolumn{1}{c}{0.16\ /0.47\ /0.63\ } \\
& SP\cite{Detone2018superpoint}+SG\cite{sarlin2020superglue} & \multicolumn{1}{c|}{0.34\ /0.67\ /0.81\ }       &  \multicolumn{1}{c}{0.21\ /0.53\ /\textbf{0.72}} \\
& SP\cite{Detone2018superpoint}+SG\cite{sarlin2020superglue}+Ada & \multicolumn{1}{c|}{0.35\ /0.71\ /0.81\ }   & \multicolumn{1}{c}{0.24\ /\textbf{0.59}\ /\textbf{0.72}} \\ \midrule
  
 & LoFTR-OT\cite{JiamingSunHujunBao2021loftr}  & \multicolumn{1}{c|}{0.41\ /0.70\ /0.79\ }   & \multicolumn{1}{c}{0.15\ /0.47\ /0.61\ } \\
 & LoFTR-DS\cite{JiamingSunHujunBao2021loftr} & \multicolumn{1}{c|}{0.44\ /\underline{0.73}\ /0.82\ }  & \multicolumn{1}{c}{0.19\ /0.54\ /0.67\ }  \\
\multirow{3}{*}{\textbf{Detector-}} & AdaMatcher-LoFTR & \multicolumn{1}{c|}{\underline{0.49}\ /\textbf{0.75}\ /\underline{0.83}} & \multicolumn{1}{c}{\underline{0.26}\ /0.57\ /0.69} \\ \cdashline{2-4}[2pt/4pt]

\multirow{3}{*}{\textbf{free}} & QuadTree\cite{tang2021quadtree} & \multicolumn{1}{c|}{0.48\ /0.70\ /0.81} & \multicolumn{1}{c}{0.20\ /0.48\ /0.65} \\
& AdaMatcher-Quad & \multicolumn{1}{c|}{0.47\ /\textbf{0.75}\ /\underline{0.83}} & \multicolumn{1}{c}{\underline{0.26}\ /\underline{0.58}\ /0.69} \\ \cdashline{2-4}[2pt/4pt]

& ASpanFormer\cite{chen2022aspanformer} & \multicolumn{1}{c|}{0.46\ /0.72\ /0.82} & \multicolumn{1}{c}{0.22\ /0.51\ /0.68} \\
& AdaMatcher-ASpan & \multicolumn{1}{c|}{\textbf{0.50}\ /\textbf{0.75}\ /\textbf{0.84}} & \multicolumn{1}{c}{\textbf{0.27}\ /0.57\ /\underline{0.70}} \\ \bottomrule
\end{tabular}}
\caption{\textbf{Homography estimation on HPatches.} 
The better methods are underlined, and the best overall method is highlighted in bold. Under viewpoint changes, AdaMatcher has substantial performance improvements compared to the corresponding baselines.}
\label{hpatch}
\end{table}

\begin{figure*}[htp]
\centering
\resizebox{.96\textwidth}{!}{
    \begin{tabular}{@{}ccccc@{}}
        \multicolumn{1}{c}{\includegraphics[]{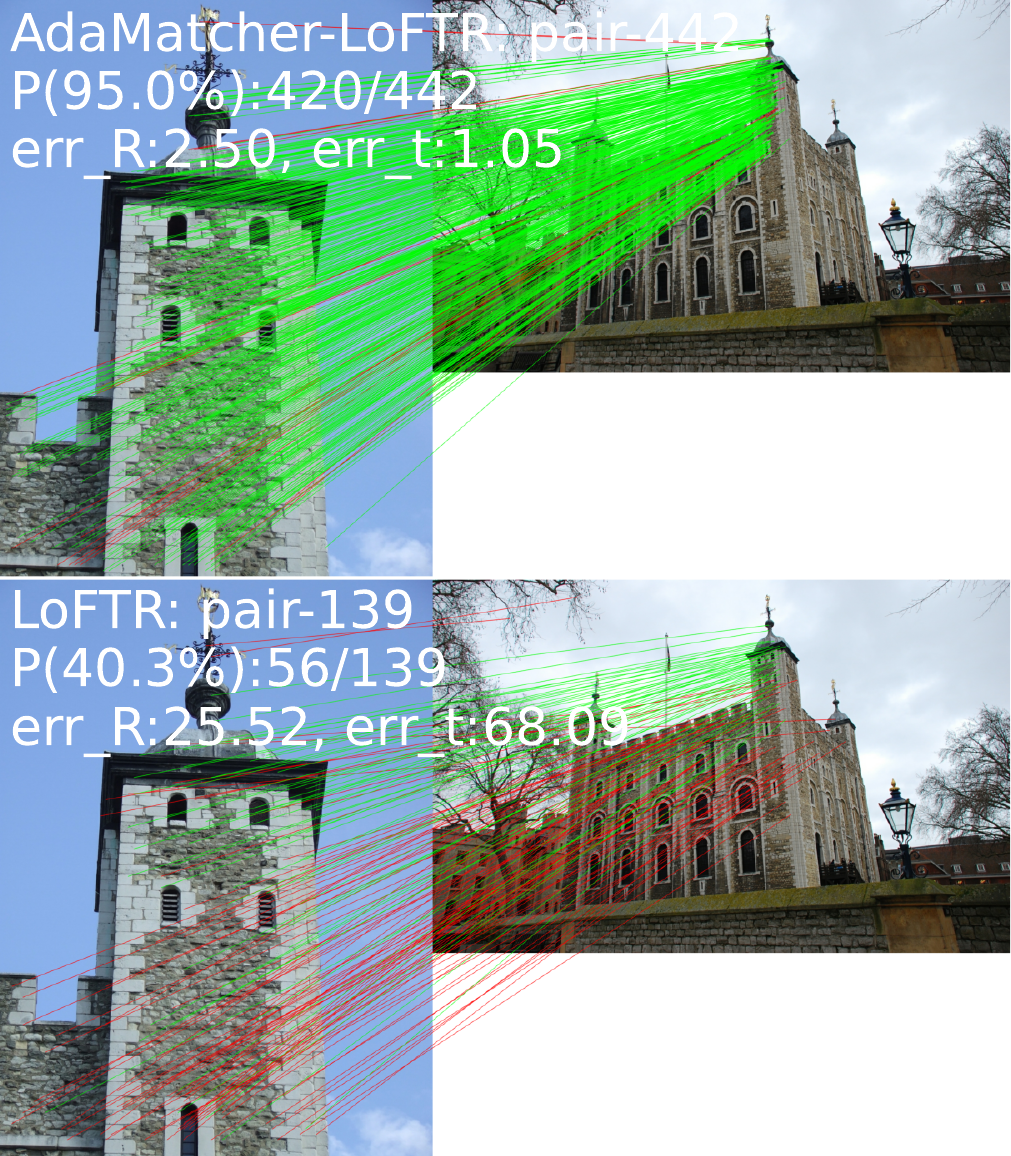}} & 
        \multicolumn{1}{c}{\includegraphics[]{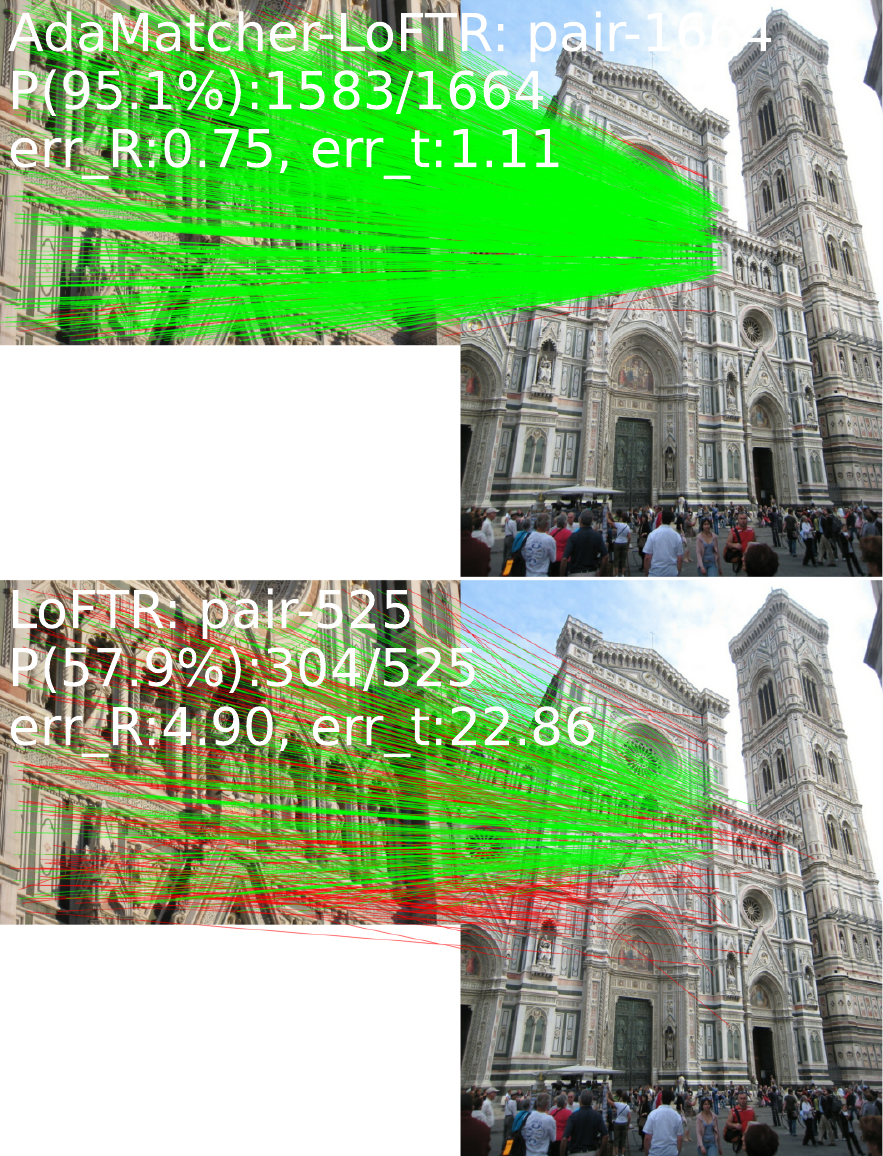}} & 
        \multicolumn{1}{c}{\includegraphics[]{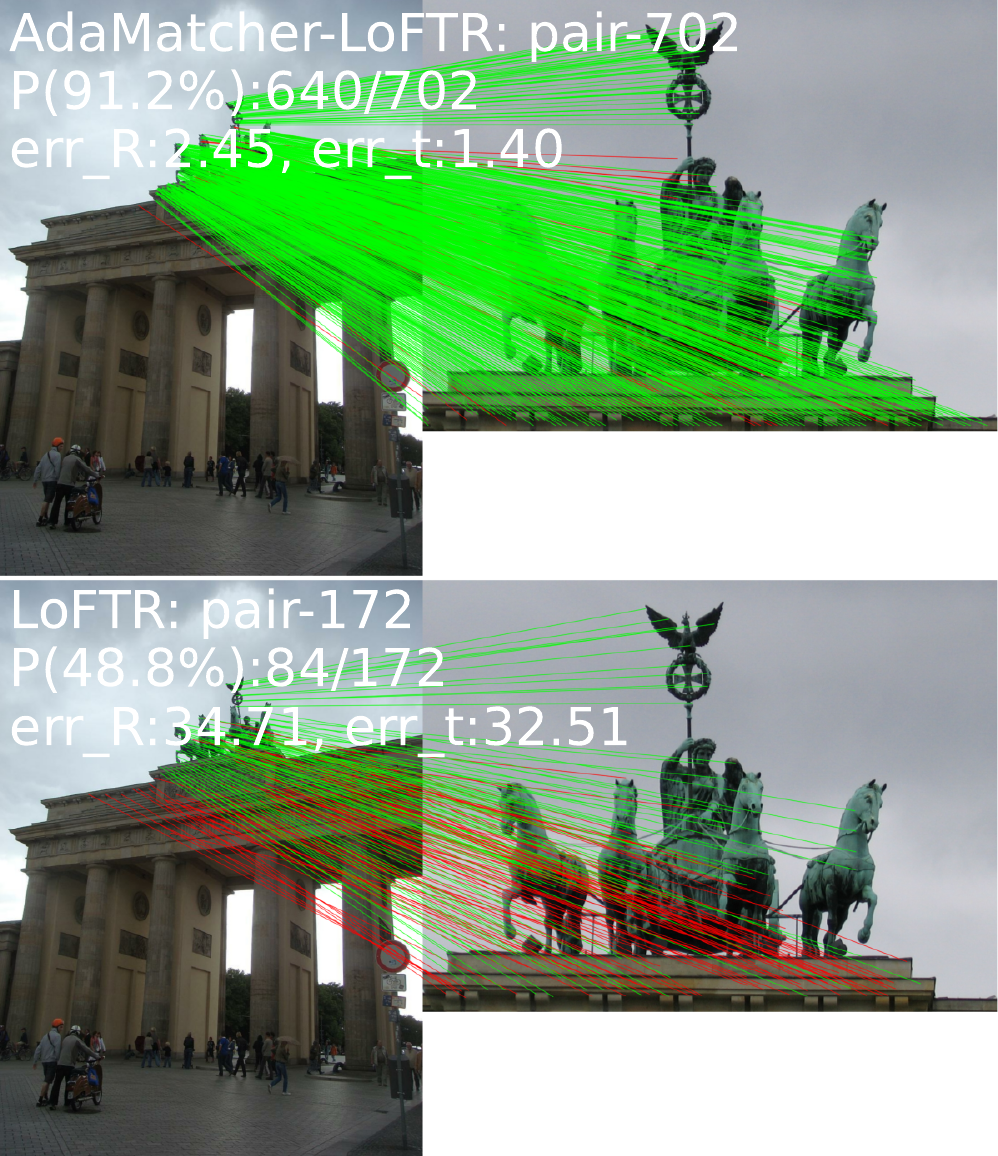}} & 
        \multicolumn{1}{c}{\includegraphics[]{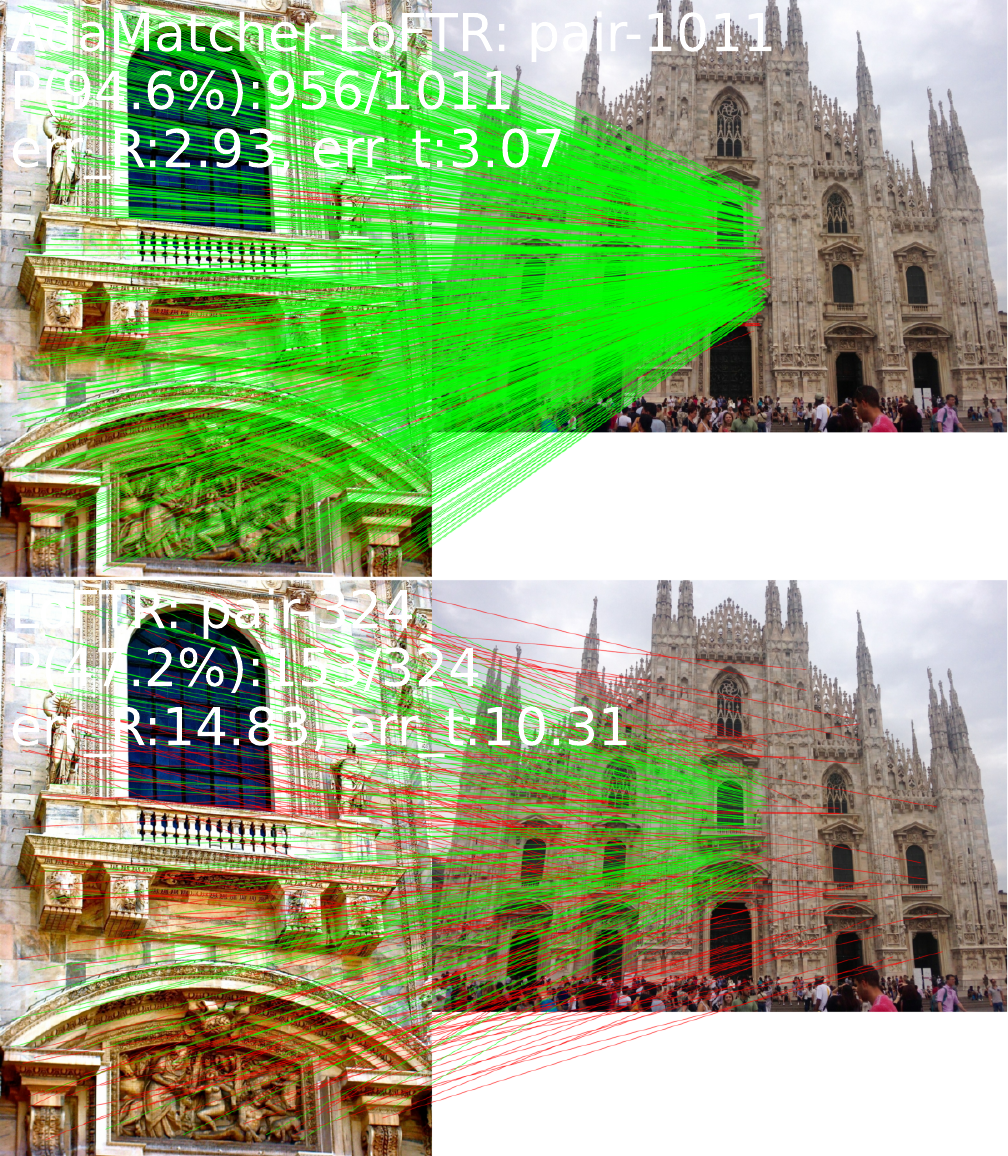}} & 
        \multicolumn{1}{c}{\includegraphics[]{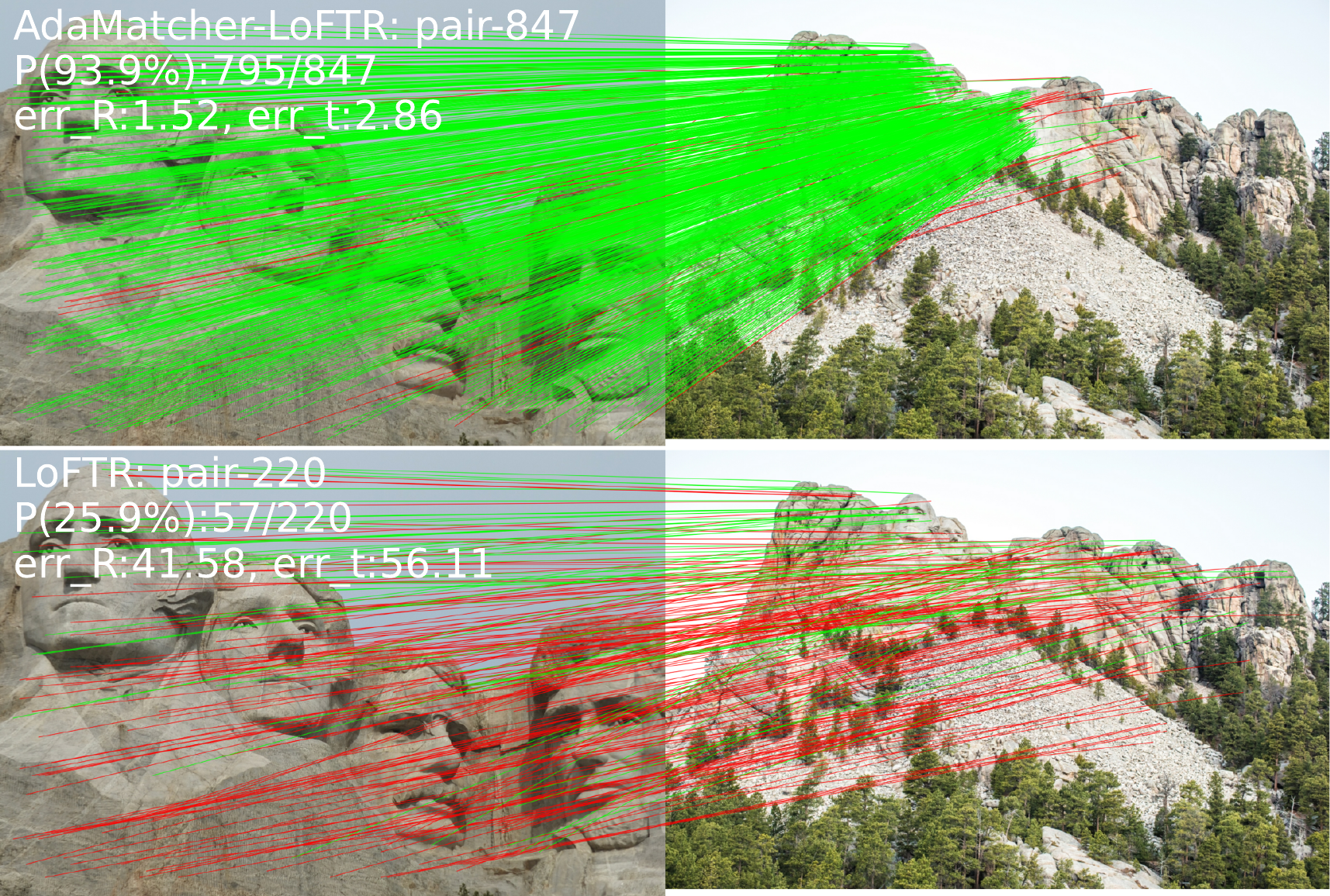}} 
    \end{tabular}
}
\caption{\textbf{Qualitative results.} AdaMatcher-LoFTR (top row) is compared to LoFTR (bottom row) in MegaDepth datasets. Matches with epipolar error beyond $1\times 10^{-4}$ are shown in green lines, and the rest are shown in red. Under scale or viewpoint variations, AdaMatcher-LoFTR performs far superior to LoFTR.}
\label{fig5}
\end{figure*}

\noindent\textbf{Results.} We split matching methods into "Detector-based" and "Detector-free" as in LoFTR\cite{JiamingSunHujunBao2021loftr}. Tab.\ref{hpatch} shows that AdaMatcher notably performs on par with or better than other baselines under all error thresholds. Under viewpoint variations, many-to-one corresponding is more appropriate, and adaptive assignment eliminates the matching ambiguity, making Adamatcher superior to other methods that use one-to-one assignment.

\subsection{Relative Pose Estimation}
\noindent\textbf{Datasets.} We use MegaDepth \cite{Li2018} to demonstrate the effectiveness of AdaMatcher for pose estimation in outdoor scenes. Following \cite{YChen2022overlap}, we used a scale-split  Megadepth test set~(with 10 scenes), as scale ratio ranges in $[1, 2)$, $ [2, 3)$, $[3, 4)$, $[4, +\infty)$. Fig.\ref{fig5} qualitatively shows the matching result of LoFTR and AdaMatcher in MegaDepth. All images (both training and test) are resized so that the longest dimension equals 840 (ASpanFormer~\cite{chen2022aspanformer} and QuadTree Attention~\cite{tang2021quadtree} use 832 due to their need for an image resolution divisible by 16). 

\noindent\textbf{Metrics.} Following \cite{sarlin2020superglue}, we report the \textbf{AUC} of the pose error under thresholds ($5^{\circ}, 10^{\circ}, 20^{\circ}$), where the pose error is set as the maximum angular error of relative rotation and translation. In our evaluation protocol, the relative poses are recovered from the essential matrix, estimated from feature matching with RANSAC. 

\noindent\textbf{Comparative methods.} We compare AdaMatcher with traditional and current SOTA methods: 1) detector-based methods including SIFT~\cite{Lowe2004sift}+HardNet~\cite{mishchuk2017working}, KeyNet+HardNet~\cite{mishchuk2017working}, R2D2~\cite{revaud2019r2d2}, ASLFeat~\cite{Luo2020aslfeat}, Disk~\cite{tyszkiewicz2020disk}, SuperGlue(SG)~\cite{sarlin2020superglue} with SuperPoint(SP)~\cite{Detone2018superpoint} or Disk~\cite{tyszkiewicz2020disk} detector and SuperGlue~\cite{sarlin2020superglue} with OETR~\cite{YChen2022overlap} for pre-processing, 2) detector-free methods including PDC-Net~\cite{truong2021pdcnet}, LoFTR~\cite{JiamingSunHujunBao2021loftr}, QuadTree Attention~\cite{tang2021quadtree} and ASpanFormer~\cite{chen2022aspanformer}. 

\begin{table*}
\centering

\resizebox{.92\textwidth}{!}{
\begin{tabular}{@{}lcccccccccccc@{}}
\toprule
\multicolumn{1}{c}{\multirow{2}{*}{\textbf{Methods}}} & \multicolumn{3}{c}{Scale {[}1,2)} &  \multicolumn{3}{c}{Scale {[}2,3)} &\multicolumn{3}{c}{Scale {[}3,4)} & \multicolumn{3}{c}{Scale {[}4,inf)} \\ \cmidrule(l){2-13} 
\multicolumn{1}{c}{} & \multicolumn{12}{c}{AUC@$5^{\circ}$ /AUC@$10^{\circ}$ /AUC@$20^{\circ}$}                             \\ \midrule

SIFT\cite{Lowe2004sift}+HardNet\cite{mishchuk2017working} & \multicolumn{1}{c}{21.19} & \multicolumn{1}{c}{33.01} & \multicolumn{1}{c|}{45.43} & \multicolumn{1}{c}{10.77} & \multicolumn{1}{c}{18.55} & \multicolumn{1}{c|}{28.64}  & \multicolumn{1}{c}{4.64} & \multicolumn{1}{c}{9.31} & \multicolumn{1}{c|}{16.21}     & \multicolumn{1}{c}{1.86} & \multicolumn{1}{c}{4.36} & \multicolumn{1}{c}{8.76} \\

KeyNet\cite{barroso2019key}+HardNet\cite{mishchuk2017working} & \multicolumn{1}{c}{34.84} & \multicolumn{1}{c}{49.08} & \multicolumn{1}{c|}{61.30} & \multicolumn{1}{c}{23.78} & \multicolumn{1}{c}{35.88} & \multicolumn{1}{c|}{47.69}  & \multicolumn{1}{c}{10.91} & \multicolumn{1}{c}{19.39} & \multicolumn{1}{c|}{29.97}     & \multicolumn{1}{c}{5.32} & \multicolumn{1}{c}{10.51} & \multicolumn{1}{c}{18.48} \\

Disk\cite{tyszkiewicz2020disk} & 33.68 & 49.76 & \multicolumn{1}{c|}{63.31} & 	5.5 & 8.45 & \multicolumn{1}{c|}{11.64} & 0.24 & 0.47 & \multicolumn{1}{c|}{0.78} & 0.09 & 0.19 & \multicolumn{1}{c}{0.35} \\

R2D2(MS)\cite{revaud2019r2d2} & \multicolumn{1}{c}{37.84} & \multicolumn{1}{c}{55.90} & \multicolumn{1}{c|}{70.66} & \multicolumn{1}{c}{22.67} & \multicolumn{1}{c}{36.93} & \multicolumn{1}{c|}{51.88}  & \multicolumn{1}{c}{6.63} & \multicolumn{1}{c}{13.02} & \multicolumn{1}{c|}{22.01}     & \multicolumn{1}{c}{2.13} & \multicolumn{1}{c}{4.02} & \multicolumn{1}{c}{7.18} \\

ASLFeat\cite{Luo2020aslfeat}   & \multicolumn{1}{c}{33.80} & \multicolumn{1}{c}{50.33} & \multicolumn{1}{c|}{65.12} & \multicolumn{1}{c}{21.87} & \multicolumn{1}{c}{35.41} & \multicolumn{1}{c|}{49.68}  & \multicolumn{1}{c}{8.53} & \multicolumn{1}{c}{16.01} & \multicolumn{1}{c|}{26.50}  & \multicolumn{1}{c}{2.95} & \multicolumn{1}{c}{6.32} & \multicolumn{1}{c}{11.84} \\

Disk\cite{tyszkiewicz2020disk}+SG\cite{sarlin2020superglue} & 45.31 & 63.04 & \multicolumn{1}{c|}{76.49} & 32.69 & 48.86 & \multicolumn{1}{c|}{63.76} & 12.38 & 22.47 & \multicolumn{1}{c|}{35.68} & 3.18 & 6.97 & \multicolumn{1}{c}{12.05} \\

SP\cite{Detone2018superpoint}+SG\cite{sarlin2020superglue}    & 
\multicolumn{1}{c}{50.43} & \multicolumn{1}{c}{67.64} & \multicolumn{1}{c|}{79.97} & \multicolumn{1}{c}{39.41} & \multicolumn{1}{c}{57.78} & \multicolumn{1}{c|}{72.34} & \multicolumn{1}{c}{19.72} & \multicolumn{1}{c}{35.22} & \multicolumn{1}{c|}{51.97} & \multicolumn{1}{c}{10.09} & \multicolumn{1}{c}{19.62} & \multicolumn{1}{c}{33.88}\\

SP\cite{Detone2018superpoint}+SG\cite{sarlin2020superglue}+Ada & \multicolumn{1}{c}{53.56} & \multicolumn{1}{c}{70.01} & \multicolumn{1}{c|}{81.90} & 
\multicolumn{1}{c}{42.32} & \multicolumn{1}{c}{59.51} & \multicolumn{1}{c|}{73.77} &
\multicolumn{1}{c}{23.77} & \multicolumn{1}{c}{39.55} & \multicolumn{1}{c|}{56.08} &
\multicolumn{1}{c}{12.63} & \multicolumn{1}{c}{23.68} & \multicolumn{1}{c}{37.59} \\

SP\cite{Detone2018superpoint}+SG\cite{sarlin2020superglue}+OETR\cite{YChen2022overlap} & \multicolumn{1}{c}{51.96} & \multicolumn{1}{c}{68.51} & \multicolumn{1}{c|}{79.95} & 
\multicolumn{1}{c}{39.92} & \multicolumn{1}{c}{56.70} & \multicolumn{1}{c|}{71.34} &
\multicolumn{1}{c}{25.37} & \multicolumn{1}{c}{41.26} & \multicolumn{1}{c|}{57.78} &
\multicolumn{1}{c}{15.36} & \multicolumn{1}{c}{28.45} & \multicolumn{1}{c}{44.27}\\
\midrule

PDC-Net(H)\cite{truong2021pdcnet} & \multicolumn{1}{c}{51.16} & \multicolumn{1}{c}{67.72} & \multicolumn{1}{c|}{79.58} & \multicolumn{1}{c}{40.35} & \multicolumn{1}{c}{56.71} & \multicolumn{1}{c|}{69.49} & \multicolumn{1}{c}{16.64} & \multicolumn{1}{c}{26.72} & \multicolumn{1}{c|}{36.77} & \multicolumn{1}{c}{4.28} & \multicolumn{1}{c}{8.14} & \multicolumn{1}{c}{12.39}\\ \hdashline[2pt/4pt]

LoFTR\cite{JiamingSunHujunBao2021loftr}  & \multicolumn{1}{c}{60.15} & \multicolumn{1}{c}{74.68} & \multicolumn{1}{c|}{84.45} & \multicolumn{1}{c}{49.69} & \multicolumn{1}{c}{65.72} & \multicolumn{1}{c|}{77.94} & \multicolumn{1}{c}{24.86} & \multicolumn{1}{c}{39.67} & \multicolumn{1}{c|}{55.08} & \multicolumn{1}{c}{10.16} & \multicolumn{1}{c}{18.74} & \multicolumn{1}{c}{29.97}\\

AdaMatcher-LoFTR & \multicolumn{1}{c}{60.50} & \multicolumn{1}{c}{74.91} & \multicolumn{1}{c|}{84.30} 
& \multicolumn{1}{c}{54.53} & \multicolumn{1}{c}{70.02} & \multicolumn{1}{c|}{81.17} 
& \multicolumn{1}{c}{35.13} & \multicolumn{1}{c}{50.75} & \multicolumn{1}{c|}{64.87} 
& \multicolumn{1}{c}{20.14} & \multicolumn{1}{c}{33.18} & \multicolumn{1}{c}{47.41} \\ \hdashline[2pt/4pt]

ASpanFormer\cite{chen2022aspanformer}  & \multicolumn{1}{c}{60.92} & \multicolumn{1}{c}{75.29} & \multicolumn{1}{c|}{85.01} & \multicolumn{1}{c}{54.60} & \multicolumn{1}{c}{70.21} & \multicolumn{1}{c|}{81.19} & \multicolumn{1}{c}{33.41} & \multicolumn{1}{c}{51.16} & \multicolumn{1}{c|}{66.88} & \multicolumn{1}{c}{18.03} & \multicolumn{1}{c}{30.50} & \multicolumn{1}{c}{44.63}\\

AdaMatcher-ASpan & \multicolumn{1}{c}{61.29} & \multicolumn{1}{c}{75.65} & \multicolumn{1}{c|}{85.41} 
& \multicolumn{1}{c}{\underline{55.35}} & \multicolumn{1}{c}{\underline{71.21}} & \multicolumn{1}{c|}{\underline{82.10}} 
& \multicolumn{1}{c}{\underline{36.05}} & \multicolumn{1}{c}{\underline{53.21}} & \multicolumn{1}{c|}{\underline{67.87}} 
& \multicolumn{1}{c}{\underline{22.92}} & \multicolumn{1}{c}{\underline{35.64}} & \multicolumn{1}{c}{\underline{50.40}} \\ \hdashline[2pt/4pt]

QuadTree\cite{tang2021quadtree}  & \multicolumn{1}{c}{\underline{62.06}} & \multicolumn{1}{c}{\textbf{76.19}} & \multicolumn{1}{c|}{\textbf{85.91}} & \multicolumn{1}{c}{53.67} & \multicolumn{1}{c}{69.83} & \multicolumn{1}{c|}{81.59} & \multicolumn{1}{c}{31.62} & \multicolumn{1}{c}{48.54} & \multicolumn{1}{c|}{64.60} & \multicolumn{1}{c}{14.77} & \multicolumn{1}{c}{26.17} & \multicolumn{1}{c}{39.89}\\

AdaMatcher-Quad & \multicolumn{1}{c}{\textbf{62.42}} & \multicolumn{1}{c}{\underline{76.03}} & \multicolumn{1}{c|}{\underline{85.42}} 
& \multicolumn{1}{c}{\textbf{56.98}} & \multicolumn{1}{c}{\textbf{71.75}} & \multicolumn{1}{c|}{\textbf{82.60}} 
& \multicolumn{1}{c}{\textbf{41.00}} & \multicolumn{1}{c}{\textbf{58.67}} & \multicolumn{1}{c|}{\textbf{73.42}} 
& \multicolumn{1}{c}{\textbf{26.56}} & \multicolumn{1}{c}{\textbf{42.05}} & \multicolumn{1}{c}{\textbf{56.71}} \\ 

\bottomrule
\end{tabular}}
\caption{\textbf{Evaluation on MegaDepth.} Performance gain from AdaMatcher becomes more prominent when scaling variation between image pairs increases. Also, our proposed method can significantly improve the performance of LoFTR and its variants.}
\label{megadepth}
\end{table*}

\noindent\textbf{Results on MegaDepth.} When the relative scale ratio is small, AdaMatcher performs slightly better than LoFTR. As the scale difference increases, AdaMatcher outperforms its counterparts more obviously. Though SIFT \cite{Lowe2004sift} (even combined with HardNet \cite{mishchuk2017working} to more discriminative descriptors) detects keypoint in scale space, R2D2 \cite{revaud2019r2d2} utilizes multi-resolution images to inference features and ASLFeat~\cite{Luo2020aslfeat} extracts features from multi-scale score maps, these methods cannot explicitly model relative scale ratio between images like AdaMatcher. 
LoFTR~\cite{JiamingSunHujunBao2021loftr} and its variants~\cite{tang2021quadtree, chen2022aspanformer} are trained using ground-truth matches obtained by one-to-one assignment, resulting in the inability to learn the geometry consistency of feature matching. 
When the scale differences between image pairs are large, the number of matches obtained by one-to-one assignment decreases, which would affect the accuracy of camera pose estimation.
Since AdaMatcher eliminates the ambiguity of matching during training, it can achieve significant improvement when inferring image pairs with large-scale variations. It can be seen that our proposed method achieves significant performance gains when applied on LoFTR~\cite{JiamingSunHujunBao2021loftr}, ASpanFormer~\cite{chen2022aspanformer} and QuadTree Attention~\cite{tang2021quadtree}.

\noindent\textbf{Refinement Module.} As mentioned before, adaptive assignment and sub-pixel refinement module could be treated as a refinement network with different extractors and matchers as Patch2Pix~\cite{zhou2021patch2pix}. Different from SuperGlue's mutual nearest neighbors constraint, we calculate row matches and column matches separately to get many-to-one and one-to-many matches, and then, refine the sub-pixel position in the descriptor feature map. As shown in Tab.\ref{megadepth}, after adding Ada as a refinement network for SP\cite{Detone2018superpoint}+SG\cite{sarlin2020superglue}, we observe a noticeable improvement in the AUC metric.

\subsection{Visual Localization}
\noindent\textbf{Datasets.} 
In HPatches and MegaDepth we only recover relative pose from feature matches. For real-world applications such as AR navigation or autonomous driving, visual localization with absolute pose estimation is a critical geometrical task. Aachen Day-Night v1.1 dataset \cite{Zhang2020} is chosen to demonstrate the visual localization ability.
 
\noindent\textbf{Experimental setup.} We use open-sourced hierarchical localization pipeline HLoc proposed in \cite{Sarlin2019} to evaluate on day-night query images. To build feature tracks for detector-free methods, we merge keypoints that are close to each other (with distance less than 4 pixels) by taking their average location, following \cite{zhou2021patch2pix}. It may not be a perfect solution with degraded sub-pixel level accuracy, but it should be a reasonable way to evaluate AdaMatcher.

\noindent\textbf{Results.}
As shown in Tab.~\ref{aachen}, AdaMatcher outperforms all the other detector-free methods. This should be attributed to the fact that adaptive assignment eliminates geometric inconsistency during training and testing. The performance of the detector-free methods is slightly lower than that of SP~\cite{Detone2018superpoint} + SG~\cite{sarlin2020superglue} on the day queries, probably due to the fact that the detector-free methods require quantification of the matches during the database reconstruction process. On the other hand, for night queries the lighting conditions are darker, making the matching process more difficult. However, with the use of adaptive assignment, the geometric consistency is increased, and the descriptive ability is improved. The improvement in matching ability compensates for the loss of quantification during the mapping process, resulting in higher performance indicators.

\begin{table}[]
\centering
\resizebox{0.48\textwidth}{!}{
\begin{tabular}{@{}l|cccccc@{}}
\toprule
\multicolumn{1}{c}{\multirow{2}{*}{Methods}} & \multicolumn{3}{c}{Day} & \multicolumn{3}{c}{Night}             \\
\multicolumn{1}{c}{}                        & \multicolumn{6}{c}{\begin{tabular}[c]{@{}c@{}}(0.25m, $2^{\circ}$) / (0.5m, $5^{\circ}$) / (1.0m, $10^{\circ}$)\end{tabular}}  \\ \midrule
SP\cite{Detone2018superpoint}+SG\cite{sarlin2020superglue} & \multicolumn{1}{c}{\textbf{89.8}} & \multicolumn{1}{c}{\textbf{96.1}} & \multicolumn{1}{c|}{\textbf{99.4}} & \multicolumn{1}{c}{77.0} & \multicolumn{1}{c}{90.6} & \multicolumn{1}{c}{\textbf{100.0}}   \\

SP\cite{Detone2018superpoint}+SG\cite{sarlin2020superglue}+Patch2Pix\cite{zhou2021patch2pix}    & \multicolumn{1}{c}{89.3}&\multicolumn{1}{c}{95.8}&\multicolumn{1}{c|}{99.2} & \multicolumn{1}{c}{78.0}&\multicolumn{1}{c}{90.6}&\multicolumn{1}{c}{99.0}    \\

Patch2Pix\cite{zhou2021patch2pix}                                      & \multicolumn{1}{c}{86.4}&\multicolumn{1}{c}{93.0}&\multicolumn{1}{c|}{97.5} & \multicolumn{1}{c}{72.3}&\multicolumn{1}{c}{88.5}&\multicolumn{1}{c}{97.9}    \\

\midrule
LoFTR-DS\cite{JiamingSunHujunBao2021loftr}  & \multicolumn{1}{c}{-}&\multicolumn{1}{c}{-}&\multicolumn{1}{c|}{-} & \multicolumn{1}{c}{72.8}&\multicolumn{1}{c}{88.5}&\multicolumn{1}{c}{99.0}    \\

LoFTR-OT\cite{JiamingSunHujunBao2021loftr}  & \multicolumn{1}{c}{88.7}&\multicolumn{1}{c}{95.6}&\multicolumn{1}{c|}{99.0} & \multicolumn{1}{c}{\underline{78.5}}&\multicolumn{1}{c}{90.6}&\multicolumn{1}{c}{99.0}    \\

ASpanFormer\cite{chen2022aspanformer} & \multicolumn{1}{c}{\underline{89.4}} & \multicolumn{1}{c}{95.6} & \multicolumn{1}{c|}{99.0} & \multicolumn{1}{c}{77.5} & \multicolumn{1}{c}{\underline{91.6}} & \multicolumn{1}{c}{\underline{99.5}}\\

AdaMatcher-LoFTR & \multicolumn{1}{c}{89.2}&\multicolumn{1}{c}{\underline{96.0}}&\multicolumn{1}{c|}{\underline{99.3}}  & \multicolumn{1}{c}{\textbf{79.1}}&\multicolumn{1}{c}{90.6}&\multicolumn{1}{c}{\underline{99.5}} \\ 

AdaMatcher-Quad & \multicolumn{1}{c}{89.2} & \multicolumn{1}{c}{95.9} & \multicolumn{1}{c|}{99.2} & \multicolumn{1}{c}{\textbf{79.1}} & \multicolumn{1}{c}{\textbf{92.1}} & \multicolumn{1}{c}{\underline{99.5}} \\
\bottomrule
\end{tabular}}
\caption{\textbf{Visual localization evaluation on the Aachen Day-Night benchmark v1.1.}}
\label{aachen}
\end{table}

\subsection{Ablation Study}
\begin{table}[htp]
\setlength\tabcolsep{1.5pt} 
\centering
\resizebox{0.48\textwidth}{!}{
\begin{tabular}{@{}ccccccccccc@{}}
\toprule
\multicolumn{3}{c}{\multirow{1}{*}{Sub Modules}} & \multicolumn{6}{c}{\multirow{1}{*}{Pose Estimation AUC}} & \multicolumn{2}{c}{\multirow{1}{*}{Precision}} \\ 
                           
\cmidrule{1-11}
\multicolumn{1}{c}{CFI}  & \multicolumn{1}{c}{AA} & \multicolumn{1}{c|}{Refine} & \multicolumn{1}{c}{@$5^{\circ}$} & \multicolumn{1}{c|}{$\Delta$}        & \multicolumn{1}{c}{@$10^{\circ}$}  & \multicolumn{1}{c|}{$\Delta$}      & \multicolumn{1}{c}{@$20^{\circ}$} & \multicolumn{1}{c|}{$\Delta$}         & \multicolumn{1}{c}{@1e-4}        & \multicolumn{1}{c}{$\Delta$}  \\ \midrule

 & & \multicolumn{1}{c|}{} &  \multicolumn{1}{c}{36.22}  & \multicolumn{1}{c|}{-} & \multicolumn{1}{c}{49.70} & \multicolumn{1}{c|}{-} & \multicolumn{1}{c}{61.86} & \multicolumn{1}{c|}{-} & \multicolumn{1}{c}{77.61} & \multicolumn{1}{c}{-}  \\

 \checkmark & & \multicolumn{1}{c|}{} & \multicolumn{1}{c}{37.13} & \multicolumn{1}{c|}{+2.5\%} & \multicolumn{1}{c}{51.17} & \multicolumn{1}{c|}{+3.0\%} & \multicolumn{1}{c}{63.56} & \multicolumn{1}{c|}{+2.7\%} & \multicolumn{1}{c}{79.13} & \multicolumn{1}{c}{+2.0\%} \\
 
 \checkmark &  & \multicolumn{1}{c|}{\checkmark}  & \multicolumn{1}{c}{38.23} & \multicolumn{1}{c|}{+5.5\%} & \multicolumn{1}{c}{52.16} & \multicolumn{1}{c|}{+4.9\%} & \multicolumn{1}{c}{64.22} & \multicolumn{1}{c|}{+3.8\%} & \multicolumn{1}{c}{78.24} & \multicolumn{1}{c}{+0.8\%} \\
 
 & \checkmark & \multicolumn{1}{c|}{} & \multicolumn{1}{c}{39.98} & \multicolumn{1}{c|}{+10.4\%} & \multicolumn{1}{c}{54.25} & \multicolumn{1}{c|}{+9.2\%} & \multicolumn{1}{c}{66.52} & \multicolumn{1}{c|}{+7.5\%} & \multicolumn{1}{c}{84.20} & \multicolumn{1}{c}{+8.5\%}\\
 
  \checkmark & \checkmark & \multicolumn{1}{c|}{} & \multicolumn{1}{c}{40.06} & \multicolumn{1}{c|}{+10.6\%} & \multicolumn{1}{c}{54.39} & \multicolumn{1}{c|}{+9.4\%} & \multicolumn{1}{c}{66.73} & \multicolumn{1}{c|}{+7.9\%} & \multicolumn{1}{c}{84.39} & \multicolumn{1}{c}{+8.7\%}\\
  
 \checkmark & \checkmark & \multicolumn{1}{c|}{\checkmark} & \multicolumn{1}{c}{\textbf{42.54}} & \multicolumn{1}{c|}{\textbf{+17.4\%}} & \multicolumn{1}{c}{\textbf{57.18}} & \multicolumn{1}{c|}{\textbf{+15.1\%}} & \multicolumn{1}{c}{\textbf{69.40}} & \multicolumn{1}{c|}{\textbf{+12.2\%}} & \multicolumn{1}{c}{\textbf{84.99}} & \multicolumn{1}{c}{\textbf{+9.5\%}} \\
 
\bottomrule
\end{tabular}
}
\caption{\textbf{Ablation of AdaMatcher.} AdaMatcher recovers more accurate relative pose compared to baseline method LoFTR and all parts are useful modules that bring noticeable performance gain for AdaMatcher.}
\label{ab}
\end{table}
To fully understand different modules in AdaMatcher and evaluate different design choices, we repeat outdoor experiments on MegaDepth with scale ranges in $[1, +\infty)$, as shown in Tab.\ref{ab}. The first row is the result of our baseline method LoFTR \cite{JiamingSunHujunBao2021loftr}, 'CFI' represents the LoFTR module (four sets of self-cross attention layers) is replaced by our Co-visible Feature Interaction (Section \ref{CFI}), 'AA' denotes replacing LoFTR's coarse-level matching with our adaptive assignment (Section \ref{m2o}), and 'Refine' denotes replacing LoFTR's fine-level matching with our sub-pixel refinement module (Section \ref{refine}). We also report match precision in normalized camera coordinates, with epipolar distance threshold of $1e^{-4}$ \cite{Detone2018superpoint, dusmanu2019d2, sarlin2020superglue}. By using CFI module, we can get more accurate matches. When the adaptive assignment is allowed in patch-level matching, the accuracy of relative pose estimation and the precision of matching are greatly improved, which means that adaptive assignment plays a vital role here. And the performance can be further enhanced by adding sub-pixel refinement module.

\subsection{Runtime Evaluation}
To test the timing of inference, we repeated the outdoor experiments on the Megadepth test set with 4000 image pairs, limiting the maximum number of matches to 1024, and the input images are resized to their longer dimensions equal to 640. As shown in Tab.\ref{inf_time}, since Adamatcher can get more high-quality matches,  its inference speed is slightly slower than LoFTR and SP+SG, but the overall execution time (matching + RANSAC) is reduced due to the improvement of inlier ratio and the matching accuracy.
\begin{table}[h]
\centering
\resizebox{0.47\textwidth}{!}{
\begin{tabular}{@{}cccccccc@{}}
\toprule
\multicolumn{1}{c}{\multirow{1}{*}{\textbf{Runtime}}} & \multicolumn{4}{c}{\multirow{1}{*}{\textbf{Adamatcher-LoFTR}}} & \multicolumn{1}{c}{\multirow{2}{*}{\textbf{LoFTR}}}
&\multicolumn{1}{c}{\multirow{2}{*}{\textbf{PDCNet}}}
&\multicolumn{1}{c}{\multirow{2}{*}{\textbf{SP+SG}}}\\\cmidrule(l){2-5} 
\multicolumn{1}{c}{(ms/pair)}& \multicolumn{1}{c}{CFI} & \multicolumn{1}{c}{AA} & \multicolumn{1}{c}{Refine} & \multicolumn{1}{c}{All}\\ 
\midrule
Matching & 23.1& 17.3& 71.6 & 157.0 & 104.9 & 577.4 & 86.2 \\
+RANSAC & - & - & - & 321.4 & 324.0 & 776.9 & 347.1 \\
\bottomrule
\end{tabular}}
\caption{\textbf{Inference time.}}
\label{inf_time}
\end{table}

\section{Conclusions}

In this paper, we find the conventional mutual nearest neighbour standard should bottleneck final performance during patch-level or pixel-level matching. The proposed AdaMatcher allows for adaptive assignment during patch-level matching, which overcomes the ambiguous underlying ground-truth label assignments, and enable estimation of the scale ratio between given image pair. We observe a noticeable performance boost, especially when the scale or viewpoint between image pairs varies. We couple co-visible feature decoding and feature interaction, enabling an additional module to be used later to obtain co-visible area.
Particularly, by plugging a dedicated sub-pixel refinement module, we can effectively achieve scale alignment and accurate sub-pixel position regression. We have conducted extensive experiments to study the effect of our findings and demonstrated the superiority of our proposed AdaMatcher. We believe that AdaMatcher will bring new insights to the feature matching community.

\clearpage
{\small
\bibliographystyle{ieee_fullname}
\bibliography{egbib}
}
\end{document}


\title{Supplementary Material $for$ \\Adaptive Assignment for Geometry Aware Local Feature Matching}

\author{Dihe Huang$^{1}$\footnotemark[1] \quad
        Ying Chen$^{2}$\footnotemark[1] \quad
        Yong Liu$^{2}$ \quad
        Jianlin Liu$^{2}$ \quad
        Shang Xu$^{2}$ \\ 
        Wenlong Wu$^{2}$ \quad
        Yikang Ding$^{1}$ \quad
        Fan Tang$^{4}$\footnotemark[2] \quad
        Chengjie Wang$^{2,3}$\footnotemark[2] \\
        $^{1}$Tsinghua Univeristy \quad $^{2}$Tencent YouTu Lab \quad $^{3}$ Shanghai Jiao Tong University\\
        $^{4}$ Institute of Computing Technology, Chinese Academy of Sciences\\
        {\tt\small \{hdh20, dyk20\}@mails.tsinghua.edu.cn \quad tfan.108@gmail.com} \\
        {\tt\small \{mumuychen, choasliu, jenningsliu, shangxu, wenlongwu, jasoncjwang\}@tencent.com}}

\maketitle
\thispagestyle{empty}
\renewcommand{\thefootnote}{\fnsymbol{footnote}}
\footnotetext[1]{These authors contributed equally.}
\footnotetext[2]{Corresponding author.}

\section{Implementation Details}
\subsection{Architecture of Co-visible Area Segmentation Module}
\begin{figure}[h]
    \centering
    \includegraphics[width=0.47\textwidth]{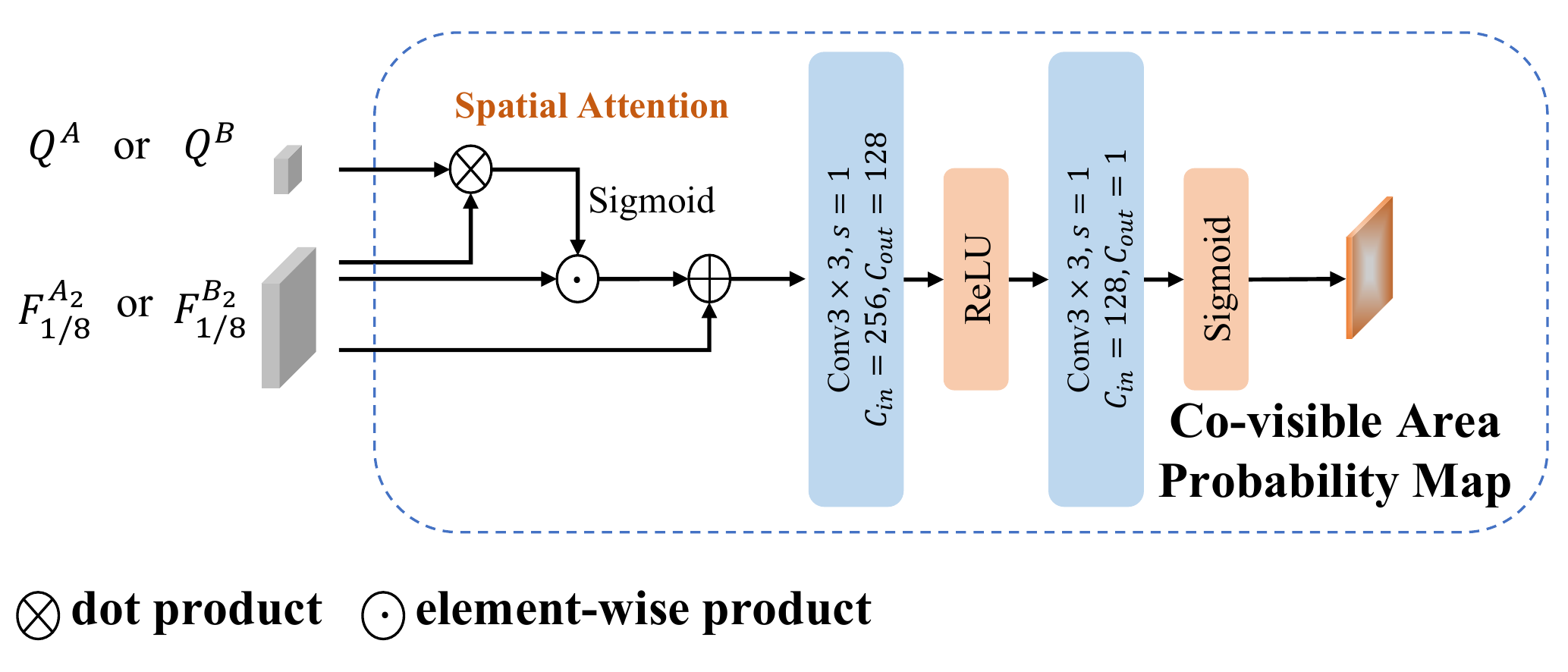}
    \centering
    \caption{\textbf{Architecture of co-visible area segmentation module.}}
    \label{sup_fig1}
\end{figure}

In order to obtain the co-visible area probability map, we borrowed the structure of DETR\cite{Carion} and used a query to perform regression on the feature map. The specific network architecture is shown in the Fig \ref{sup_fig1}. Firstly, the spatial attention map $F_{attn}^{i} \in \mathbb{R}^{1\times h\times w}$ by the dot product operation of $Q^{i}\in \mathbb{R}^{1\times 1\times C}$ and $F_{1/8}^{{i}_2}\in \mathbb{R}^{C\times h\times w}$, and then perform element-wise multiplication of $F_{attn}^{i}$ and $F_{1/8}^{{i}_2}$, followed by a shortcut connection to obtain $F_{co}^{i}\in \mathbb{R}^{C\times h\times w}$. Finally, a simple block with two convolution layers are used to obtain the co-visible area probability map, where the first convolution is followed by a ReLU activation and the second convolution is followed by a Sigmoid function. 

\subsection{Refine Network for SuperGlue}
As stated in the main text, our adaptive assignment and sub-pixel refine module can be treated as a refinement network for other matching methods, such as SuperGlue\cite{sarlin2020superglue}. Here, we will present in detail how to use our method as a refinement network for SuperGlue. To perform adaptive assignment, we simply remove the mutual nearest neighbours constraint, and obtain proposal matches by applying $argmax$ operation on each of the two dimensions of the matching matrix generated from SuperGlue, and then filtering by adding a confidence threshold. The matching matrix can be the result of multiple iterations of the skinhorn algorithm, or the matrix obtained by decomposing the last iteration of the skinhorn algorithm. After obtaining the proposal matches, we sample the patch features $F^{A}, F^{B}\in \mathbb{R}^{n\times c\times w \times w }$ at the corresponding positions on the descriptor feature maps generated by SuperPoint~\cite{Detone2018superpoint} and feed them into our one-to-one refine module (Section 3.3 of the main text) to achieve scale alignment and sub-pixel location regression. 

\section{More Experiments}
\subsection{Additional Evaluation Metrics for HPatches}
\noindent\textbf{Metrics.}
Since the homography estimation accuracy contains the effect of the OpenCV-RANSAC, we use Mean Matching Accuracy (MMA) on the Hpatches \cite{Balntas2017} dataset to evaluate different methods. We use the ratio of correctly matched features within thresholds of 1, 2, and 3 pixels, respectively, and the maximum amount of matches is limited to 1024.
\noindent\textbf{Results} in Tab.\ref{hpatch_mma} show that Adamatcher outperforms other methods in terms of matching accuracy. Since adaptive assignment eliminates the ambiguity of matching in supervision and inference, AdaMatcher is able to generate more accurate matches when the viewing angle changes.
\begin{table}[ht]
    \centering
    \resizebox{.47\textwidth}{!}{
    \begin{tabular}{@{}lllllll@{}}
    \toprule
    \multicolumn{1}{c}{\multirow{2}{*}{\textbf{Methods}}} & \multicolumn{3}{|c}{\multirow{1}{*}{\textbf{All}}} & \multicolumn{3}{|c}{\multirow{1}{*}{\textbf{Viewpoint}}} \\
         & \multicolumn{3}{|c}{MMA@1px / 2px / 3px} & \multicolumn{3}{|c}{MMA@1px / 2px / 3px}  \\
    \midrule
    SIFT\cite{Lowe2004sift}+HardNet\cite{mishchuk2017working} & \multicolumn{3}{|c}{0.460 / 0.718 / 0.828} & 
    \multicolumn{3}{|c}{0.383 / 0.671 / 0.800} \\
    KeyNet\cite{barroso2019key}+HardNet\cite{mishchuk2017working} & \multicolumn{3}{|c}{0.422 / 0.701 / 0.837} & 
    \multicolumn{3}{|c}{0.342 / 0.661 / 0.811} \\
    R2D2\cite{revaud2019r2d2} & \multicolumn{3}{|c}{0.334 / 0.611 / 0.751} & 
    \multicolumn{3}{|c}{0.273 / 0.566 / 0.699} \\
    SP\cite{Detone2018superpoint}+SG\cite{sarlin2020superglue} & \multicolumn{3}{|c}{0.367 / 0.685 / 0.827} & 
    \multicolumn{3}{|c}{0.315 / 0.654 / 0.815} \\
    SP\cite{Detone2018superpoint}+SG\cite{sarlin2020superglue}+Ada & \multicolumn{3}{|c}{\textbf{0.386} / \textbf{0.702} / \textbf{0.839}} & 
    \multicolumn{3}{|c}{\textbf{0.348} / \textbf{0.685} / \textbf{0.830}} \\ \midrule
    LoFTR-OT\cite{JiamingSunHujunBao2021loftr} & \multicolumn{3}{|c}{0.593 / 0.814 / 0.893} & 
    \multicolumn{3}{|c}{0.461 / 0.731 / 0.836} \\
    LoFTR-DS\cite{JiamingSunHujunBao2021loftr} & \multicolumn{3}{|c}{0.613 / 0.830 / 0.902} & 
    \multicolumn{3}{|c}{0.495 / 0.759 / 0.853} \\
    AdaMatcher-LoFTR & \multicolumn{3}{|c}{0.628 / 0.845 / 0.914} & 
    \multicolumn{3}{|c}{0.540 / 0.798 / 0.882} \\ \hdashline[2pt/4pt]
    QuadTree\cite{tang2021quadtree} & \multicolumn{3}{|c}{0.642 / 0.844 / 0.911} & \multicolumn{3}{|c}{0.517 / 0.777 / 0.870}\\
    AdaMatcher-QuadTree & \multicolumn{3}{|c}{0.640 / 0.850 / 0.917} & \multicolumn{3}{|c}{0.550 / 0.799 / 0.881} \\ \hdashline[2pt/4pt]
    ASpanFormer\cite{chen2022aspanformer} & \multicolumn{3}{|c}{\textbf{0.658} / 0.856 / 0.920} & \multicolumn{3}{|c}{0.539 / 0.795 / 0.881} \\
    AdaMatcher-ASpan & \multicolumn{3}{|c}{0.657 / \textbf{0.857} / \textbf{0.920}} & \multicolumn{3}{|c}{\textbf{0.552} / \textbf{0.802} / \textbf{0.887}} \\
    \bottomrule
    \end{tabular}}
    \caption{MMA metrics on HPatches.}
    \label{hpatch_mma}
\end{table}

\subsection{Results on YFCC100M}\label{result_yfcc}
The YFCC100M~\cite{thomee2016yfcc100m} dataset is also used to conduct experiments to compare AdaMatcher with several baseline methods.  To be fair, We use the same test pairs (a total of 4000 pairs) as in previous works~\cite{sarlin2020superglue, oanet2019iccv}, using their evaluation metrics. The test set is derived from four selected landmark sequences, each sampling 1000 image pairs. All images are resized to 480 $\times$ 640 and all models are trained in MegaDepth~\cite{Li2018}. The accuracy of pose estimation is measured by AUC under error thresholds ($5^\circ$, $10^\circ$ and $20^\circ$). The results are shown in Tab.\ref{yfcc100m}, where our methods all perform better than the corresponding baseline methods.

\begin{table}[ht]
    \centering
    \begin{tabular}{@{}lccc@{}}
    \toprule
    \multicolumn{1}{l}{\multirow{2}{*}{\textbf{Methods}}} & \multicolumn{3}{c}{\textbf{Pose Estimation AUC}} \\ \cmidrule{2-4}
    & @$5^{\circ}$ & @$10^{\circ}$ & @$20^{\circ}$ \\ \midrule
    LoFTR-DS~\cite{JiamingSunHujunBao2021loftr} & 43.06 & 62.21 & 77.26 \\
    AdaMatcher-LoFTR & \textbf{44.06} & \textbf{63.04} & \textbf{77.61} \\ \hdashline[2pt/4pt]
    ASpanFormer~\cite{chen2022aspanformer} & 43.70 & 62.57 & 77.32 \\
    AdaMatcher-ASpan & \textbf{43.93} & \textbf{62.92} & \textbf{77.54} \\ \hdashline[2pt/4pt]
    QuadTree~\cite{tang2021quadtree} & 36.50 & 55.64 & 71.61 \\
    AdaMatcher-Quad & \textbf{44.20} & \textbf{63.09} & \textbf{77.59} \\
    \bottomrule
    \end{tabular}
    \caption{The results of outdoor relative pose estimation on YFCC100M.}
    \label{yfcc100m}
\end{table}

\subsection{Indoor Pose Estimation}
To validate the generalizability of different detector-free methods, we perform indoor pose estimation experiments on the ScanNet~\cite{scannet2017cvpr} dataset using models trained on the MegaDepth~\cite{Li2018} dataset. We use the test split with 1500 image pairs following the experimental setting of \cite{sarlin2020superglue, JiamingSunHujunBao2021loftr, chen2022aspanformer}. To align with the existing methods \cite{JiamingSunHujunBao2021loftr, chen2022aspanformer, tang2021quadtree}, we resized all test images to $480\times640$. We use the same evaluation protocols as in Sec.~\ref{result_yfcc}. As presented in Tab.\ref{scannet}, AdaMatcher has a significant performance improvement on different baselines~\cite{JiamingSunHujunBao2021loftr, chen2022aspanformer, tang2021quadtree}. 

\begin{table}[ht]
    \centering
    \begin{tabular}{@{}lccc@{}}
    \toprule
    \multicolumn{1}{l}{\multirow{2}{*}{\textbf{Methods}}} & \multicolumn{3}{c}{\textbf{Pose Estimation AUC}} \\ \cmidrule{2-4}
    & @$5^{\circ}$ & @$10^{\circ}$ & @$20^{\circ}$ \\ \midrule
    LoFTR-OT~\cite{JiamingSunHujunBao2021loftr} & 15.46 & 31.28 & 47.87 \\
    LoFTR-DS~\cite{JiamingSunHujunBao2021loftr} & 17.26 & 33.93 & 50.16 \\
    AdaMatcher-LoFTR & \textbf{18.60} & \textbf{35.00} & \textbf{50.75} \\ \hdashline[2pt/4pt]
    ASpanFormer~\cite{chen2022aspanformer} & 20.64 & 39.34 & 56.61 \\
    AdaMatcher-ASpan & \textbf{21.33} & \textbf{39.93} & \textbf{56.69} \\ \hdashline[2pt/4pt]
    QuadTree~\cite{tang2021quadtree} & 19.83 & 37.86 & 55.03 \\
    AdaMatcher-Quad & \textbf{21.18} & \textbf{39.71} & \textbf{56.22} \\
    \bottomrule
    
    \end{tabular}
    \caption{The results of indoor relative pose estimation on ScanNet. All models are trained on MegaDepth dataset.}
    \label{scannet}
\end{table}

\subsection{Computational Costs of Feature Interaction}
We evaluate the computation and parameters between LoFTR's feature interaction module\cite{JiamingSunHujunBao2021loftr} and our CFI module (using linear attention~\cite{lineartransformer} as in LoFTR). The size of input tensor is $60\times 80 \times 256$. As shown in Tab.\ref{param}, compared to LoFTR's feature interaction module (consisting of four sets of self- and cross-attention layers), our CFI module reduces about $38.79\%$ of the computational costs and $14.29\%$ of the parameters.

\begin{table}[h]
\centering
\begin{tabular}{@{}c|cc@{}}
\toprule
\multicolumn{1}{c}{\multirow{1}{*}{Method}}  & \multicolumn{1}{c}{Flops(G)} & \multicolumn{1}{c}{Param(MB)} \\ 
\midrule
LoFTR module  & 51.82 & 5.25 \\  
CFI & 31.74 & 4.50 \\
\bottomrule
\end{tabular}

\caption{Computational complexity of feature interaction module}
\label{param}
\end{table}
\begin{figure}[ht]
    \centering
        \centering
        \rotatebox[]{0}{Projection error threshold of 3 pixel}\\
        
        \includegraphics[width=0.47\textwidth]{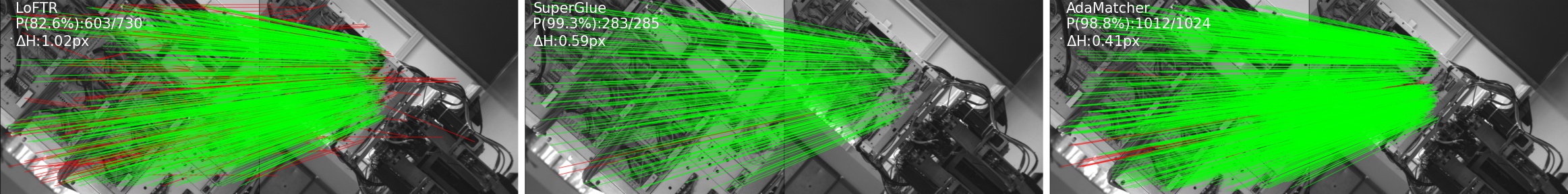}\\
        \includegraphics[width=0.47\textwidth]{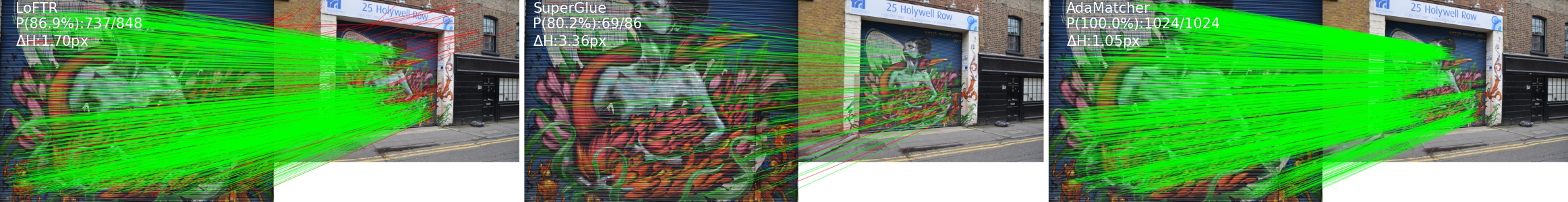}\\
        \includegraphics[width=0.47\textwidth]{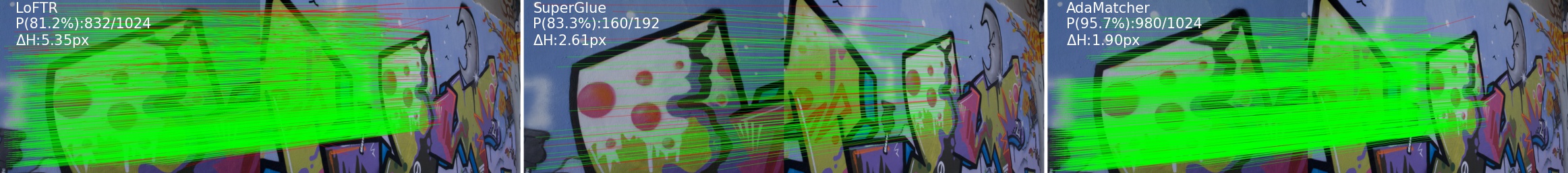}\\
        \rotatebox[]{0}{Projection error threshold of 1 pixel}\\
        \includegraphics[width=0.47\textwidth]{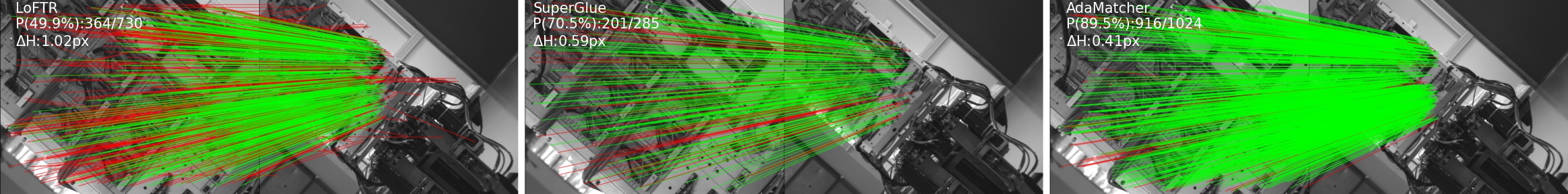}\\
        \includegraphics[width=0.47\textwidth]{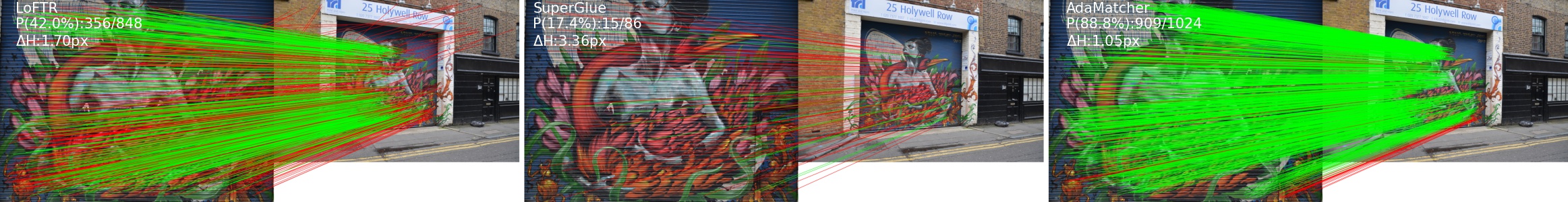}\\
        \includegraphics[width=0.47\textwidth]{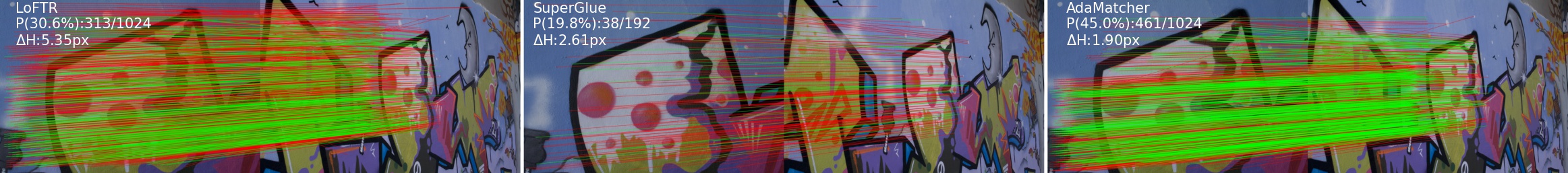}\\
        
    
    \begin{tabular}{@{}lcr@{}}
        \multicolumn{1}{l}{LoFTR\ \ \ \ \ \ \ \ } & \multicolumn{1}{c}{\ \ \ \ \ \ \ \ SP+SG\ \ \ \ \ \ \ \ } & \multicolumn{1}{r}{\ \ \ AdaMatcher-LoFTR}\\
    \end{tabular}
    \centering
    \caption{Qualitative image matches on Hpatches dataset. Matches with projection error less than the threshold are displayed in green, otherwise they are displayed in red.}
    \label{hpatch_figure}
\end{figure}

\section{D. Qualitative Results}
\begin{figure*}[ht]
    \resizebox{.82\textwidth}{!}{
    \begin{tabular}{@{}cccccccccc@{}}
        \multicolumn{1}{c}{} & \multicolumn{3}{c}{SuperPoint+SuperGlue} & \multicolumn{3}{c}{\ \ \ \ \ \ \ \ \ \ \ \ \ \ \ \ \ \ \ \ \ \ LoFTR} & \multicolumn{3}{c}{\ \ \ \ \ \ \ \ \ \ \ \ AdaMatcher-LoFTR}\\
        \multicolumn{1}{c|}{\multirow{6}{*}{\rotatebox{90}{Easy}}} 
        & \multicolumn{9}{c}{\includegraphics[width=0.9\linewidth]{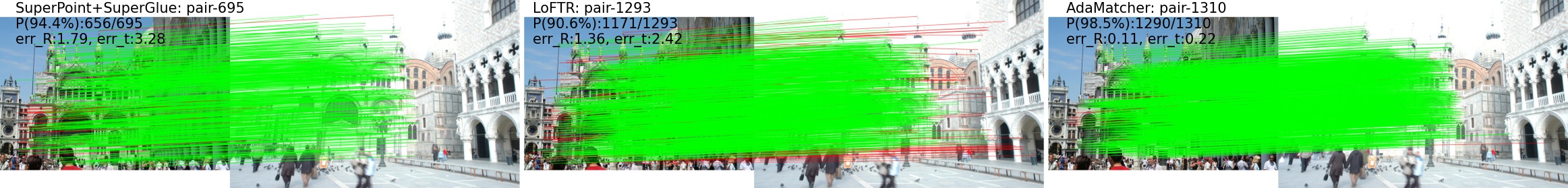}}\\   
        \multicolumn{1}{c|}{} & \multicolumn{9}{c}{\includegraphics[width=0.9\linewidth]{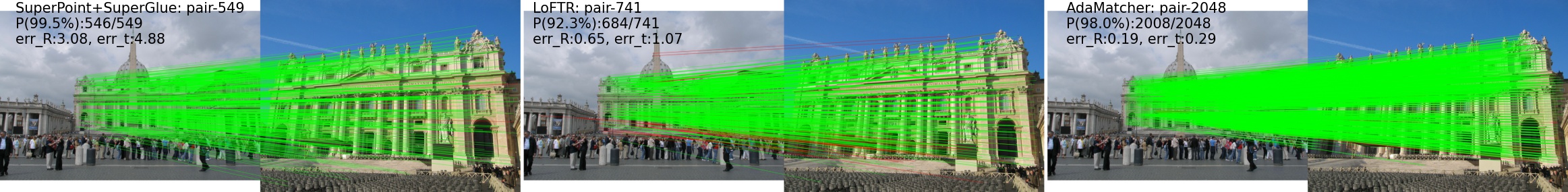}}\\     
        \multicolumn{1}{c|}{} & \multicolumn{9}{c}{\includegraphics[width=0.9\linewidth]{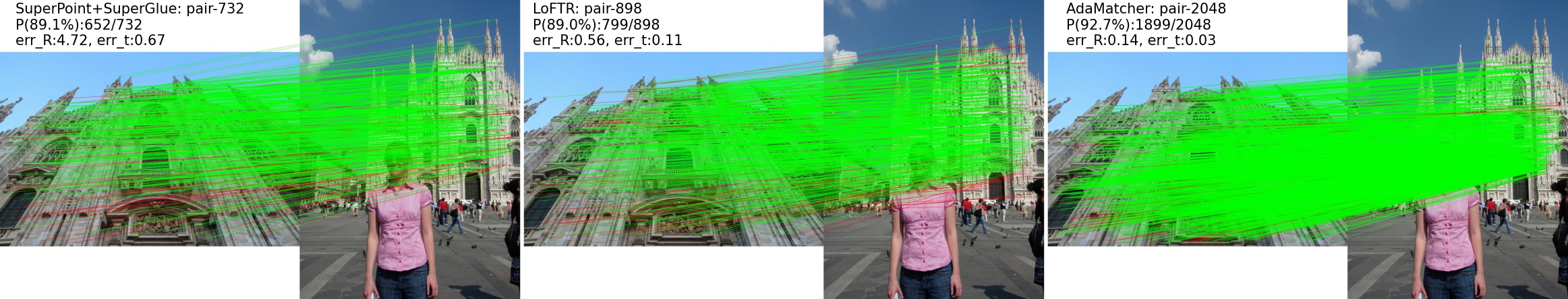}}\\     
        \cmidrule{2-10}
        
        \multicolumn{1}{c|}{\multirow{6}{*}{\rotatebox{90}{Moderate}}} 
        & \multicolumn{9}{c}{\includegraphics[width=0.9\linewidth]{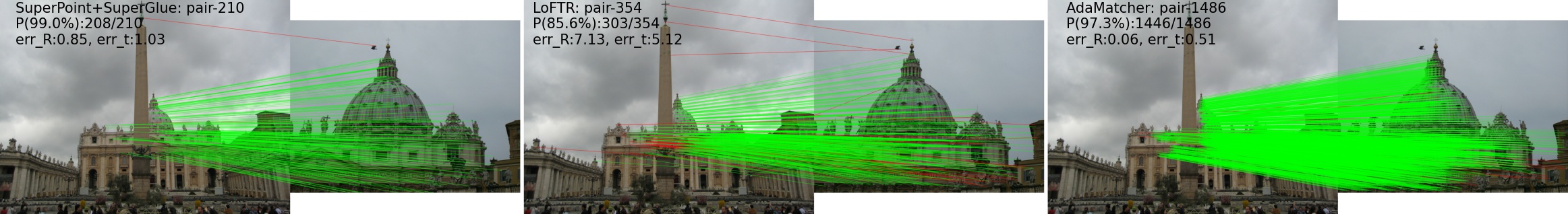}}\\ 
        \multicolumn{1}{c|}{} & \multicolumn{9}{c}{\includegraphics[width=0.9\linewidth]{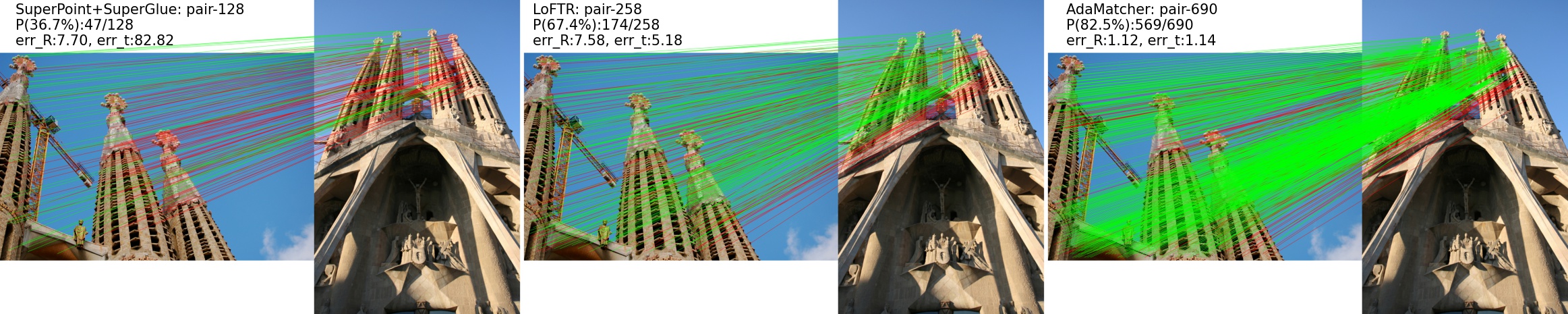}}\\
        \multicolumn{1}{c|}{} & \multicolumn{9}{c}{\includegraphics[width=0.9\linewidth]{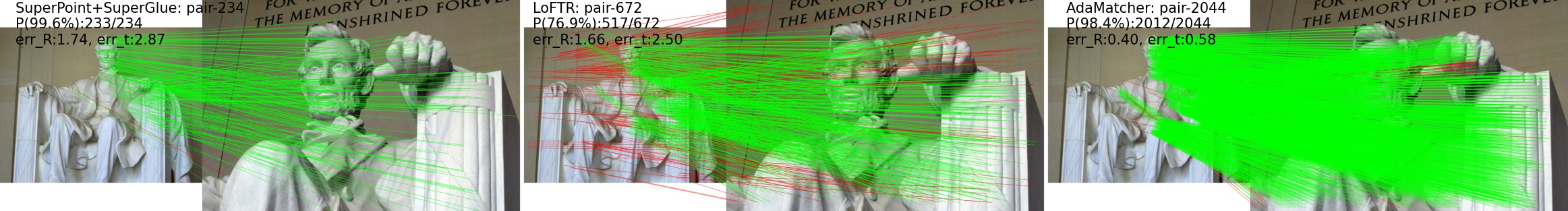}}\\
        \cmidrule{2-10}
        
        \multicolumn{1}{c|}{\multirow{14}{*}{\rotatebox{90}{Difficult}}} 
        & \multicolumn{9}{c}{\includegraphics[width=0.9\linewidth]{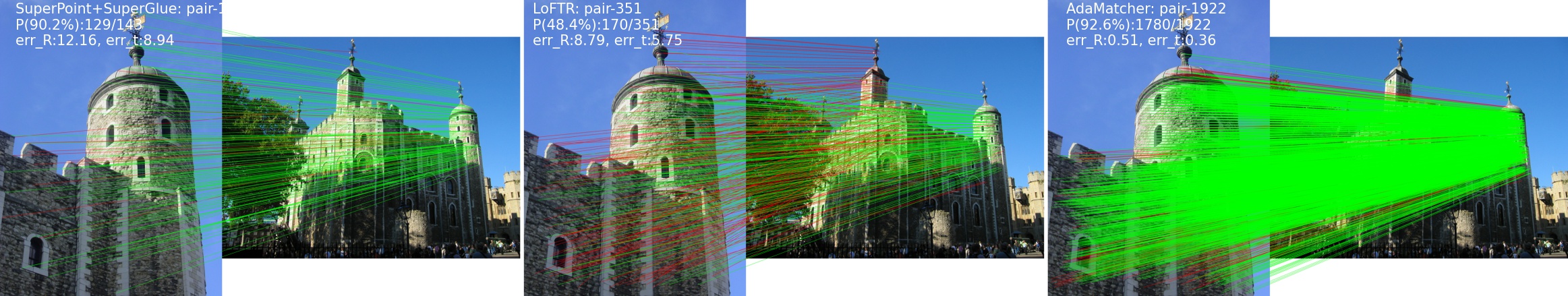}}\\ 
        \multicolumn{1}{c|}{} & \multicolumn{9}{c}{\includegraphics[width=0.9\linewidth]{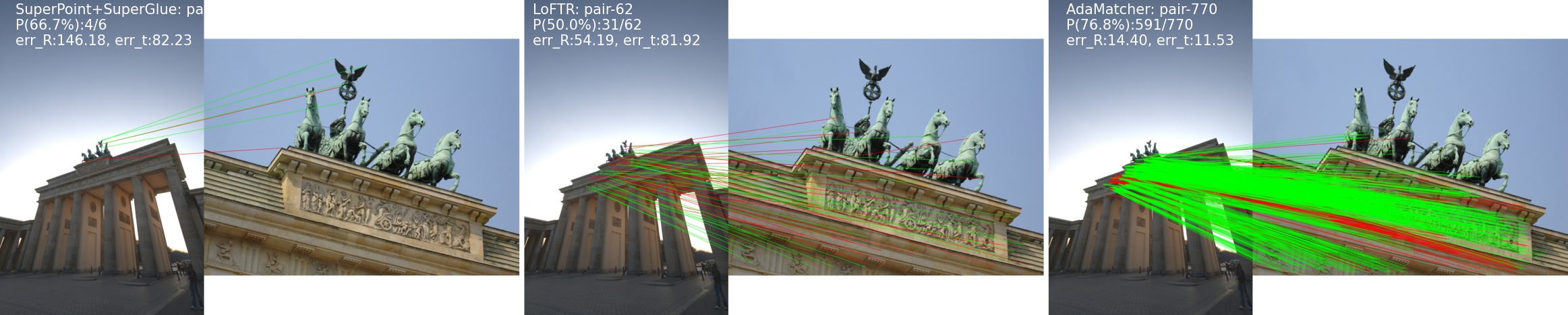}}\\
        \multicolumn{1}{c|}{} & \multicolumn{9}{c}{\includegraphics[width=0.9\linewidth]{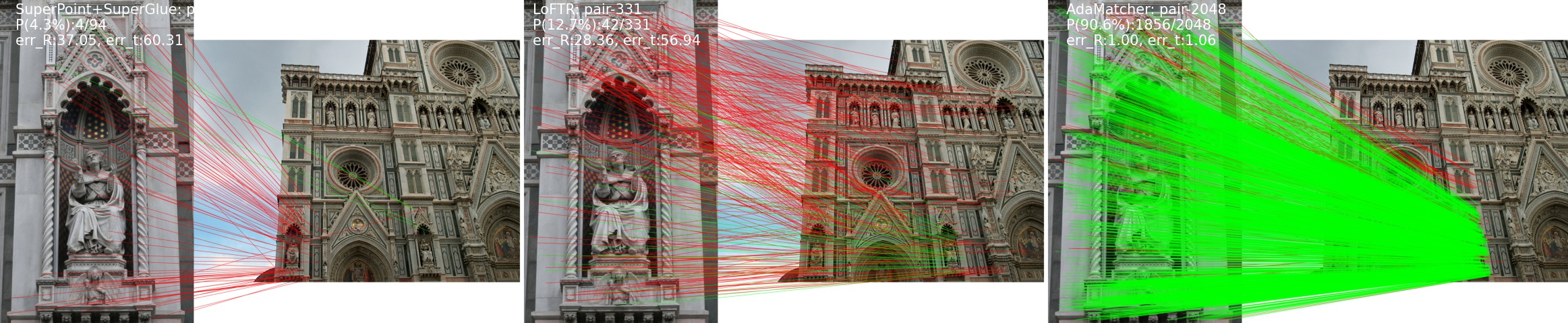}}
    \end{tabular}
    }
    \centering
    \caption{Qualitative image matches on MegaDepth dataset. Green indicates that epipolar error in normalized image coordinates is less than $1\times 10^{-4}$, while red indicates that it is exceeded.}
    \label{qualitative}
   
\end{figure*}

\begin{figure*}[ht]
    \centering
    \begin{minipage}[b]{0.32\linewidth}
        \fbox{\includegraphics[width=0.9\linewidth]{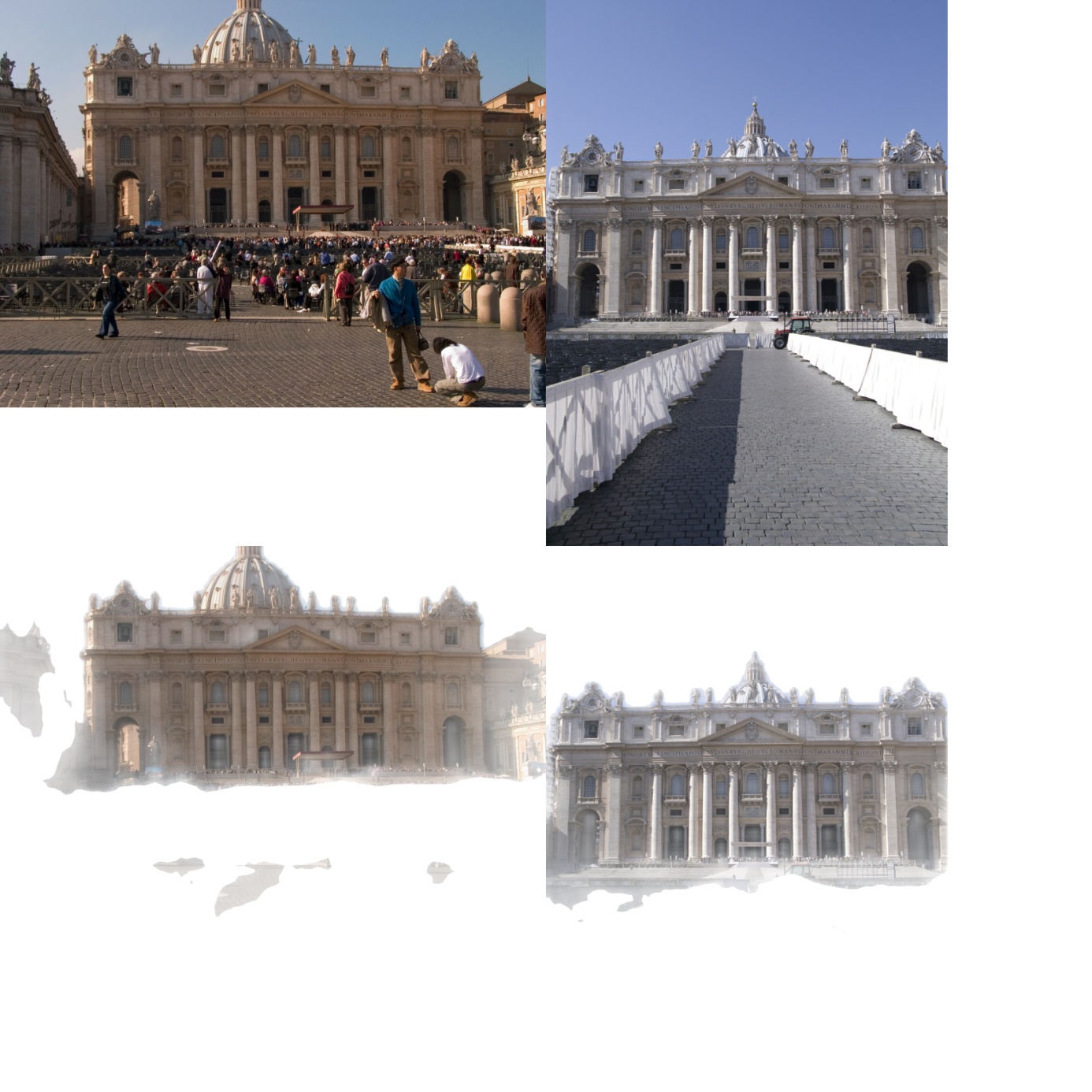}}\\
        \fbox{\includegraphics[width=0.9\linewidth]{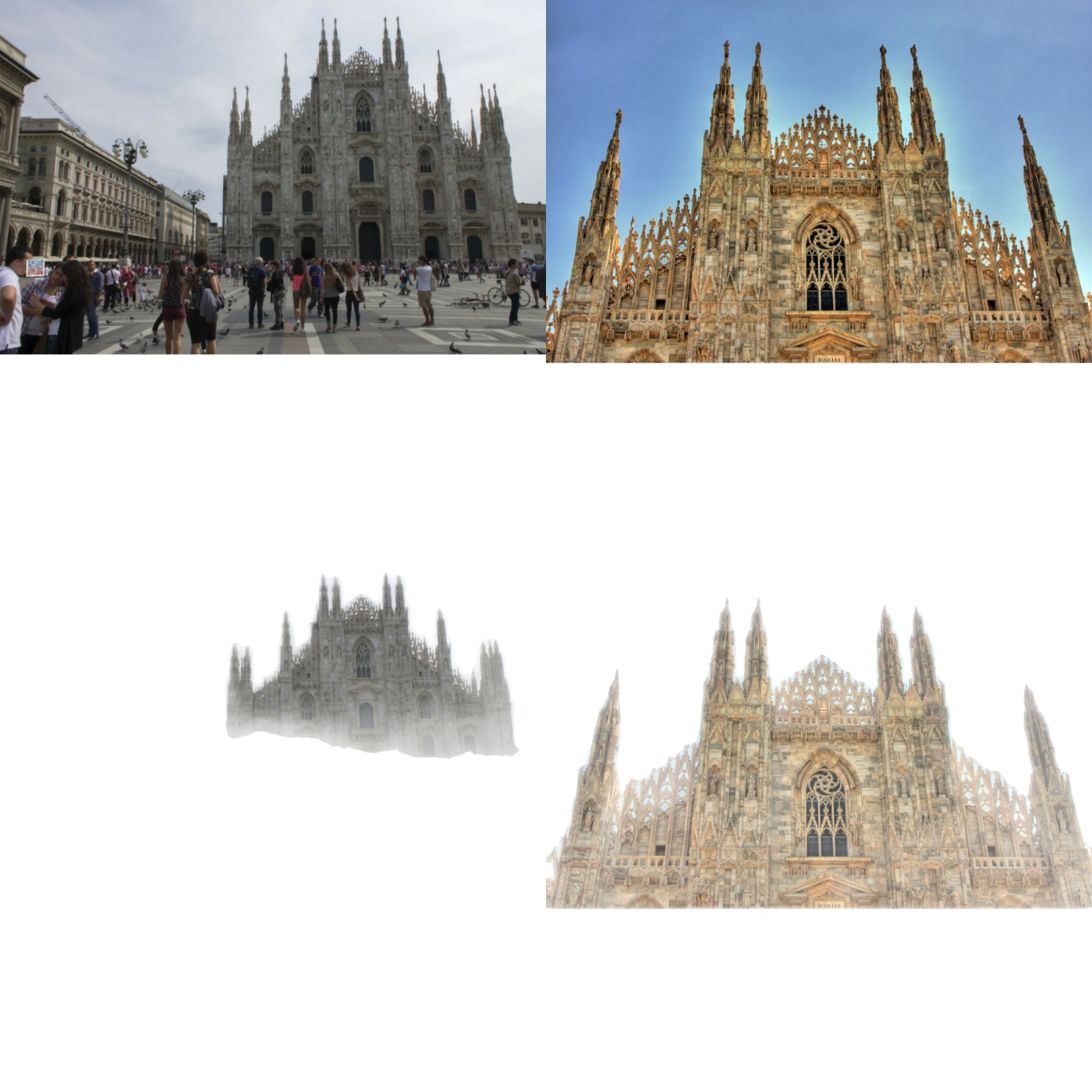}}\\  
        \fbox{\includegraphics[width=0.9\linewidth]{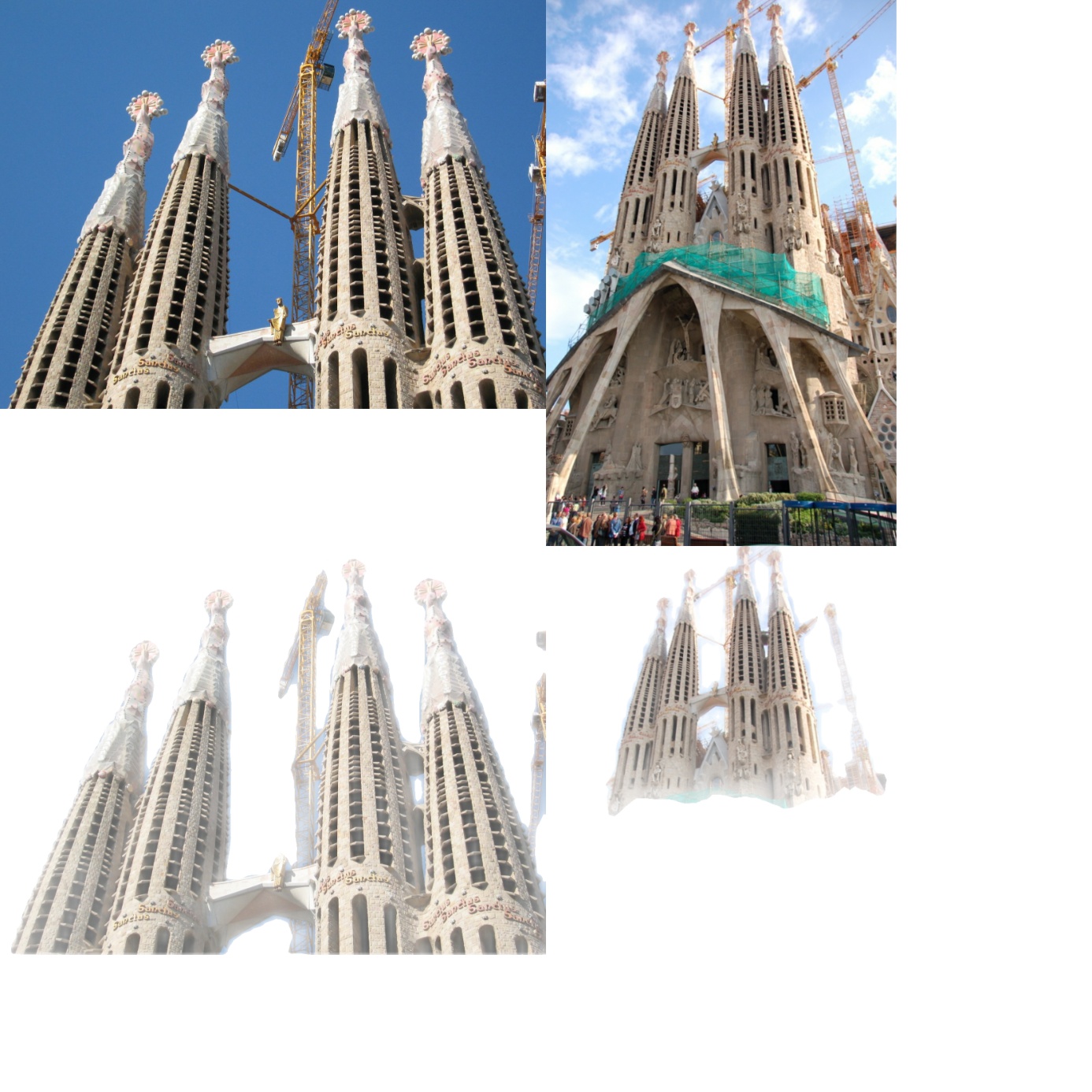}}\\
        \fbox{\includegraphics[width=0.9\linewidth]{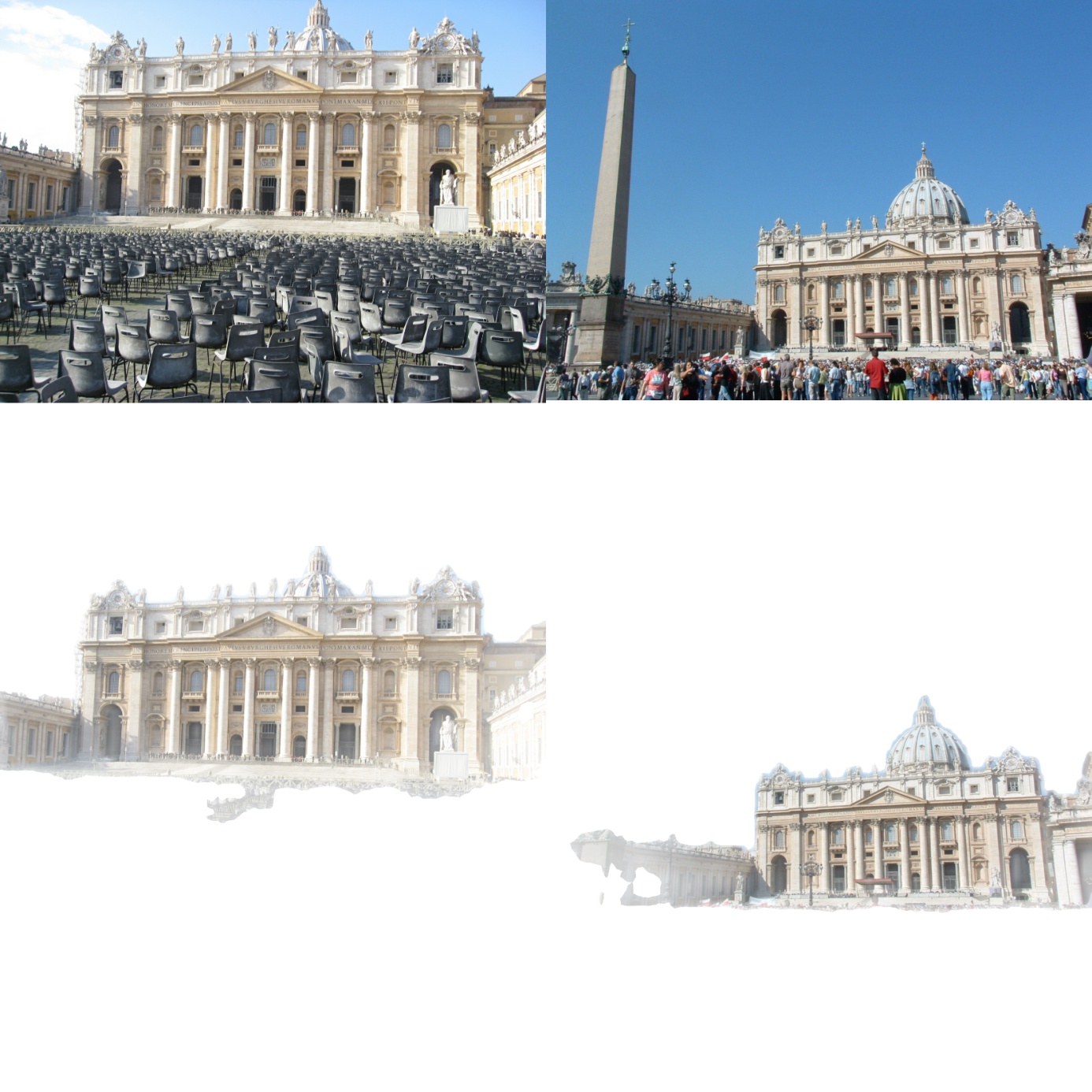}}
        
    \end{minipage}
    \begin{minipage}[b]{0.32\linewidth}
    \fbox{\includegraphics[width=0.9\linewidth]{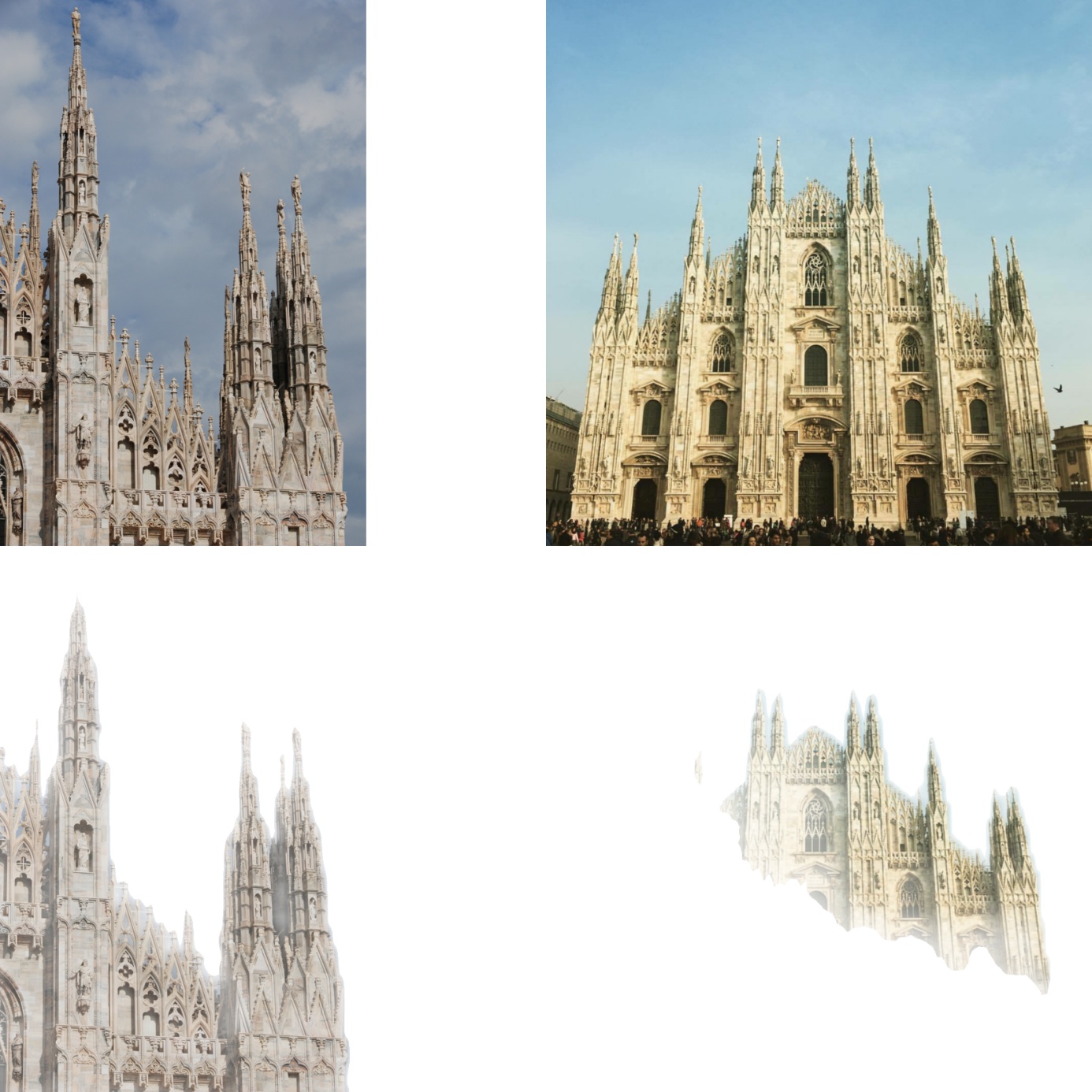}}\\
        \fbox{\includegraphics[width=0.9\linewidth]{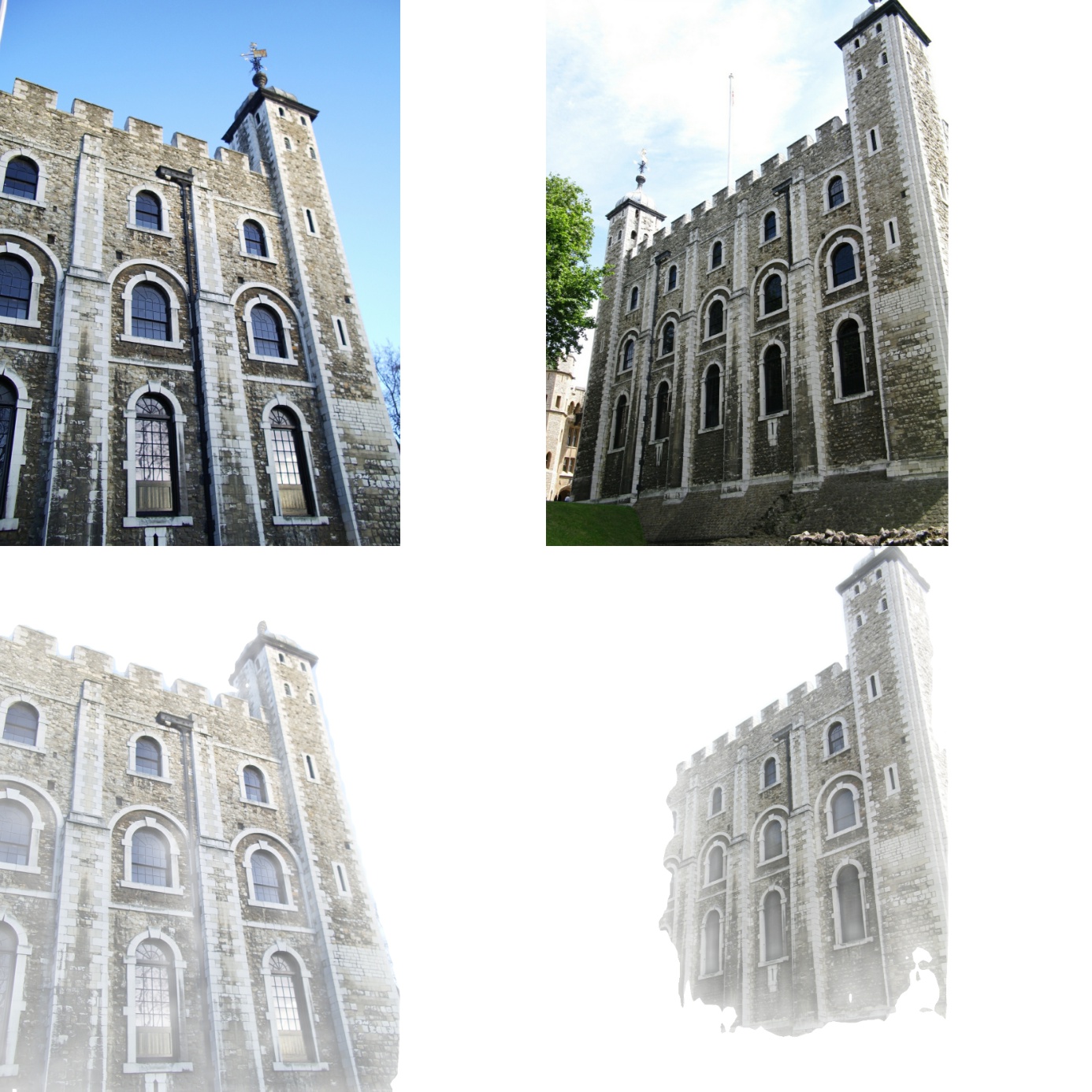}}\\
        \fbox{\includegraphics[width=0.9\linewidth]{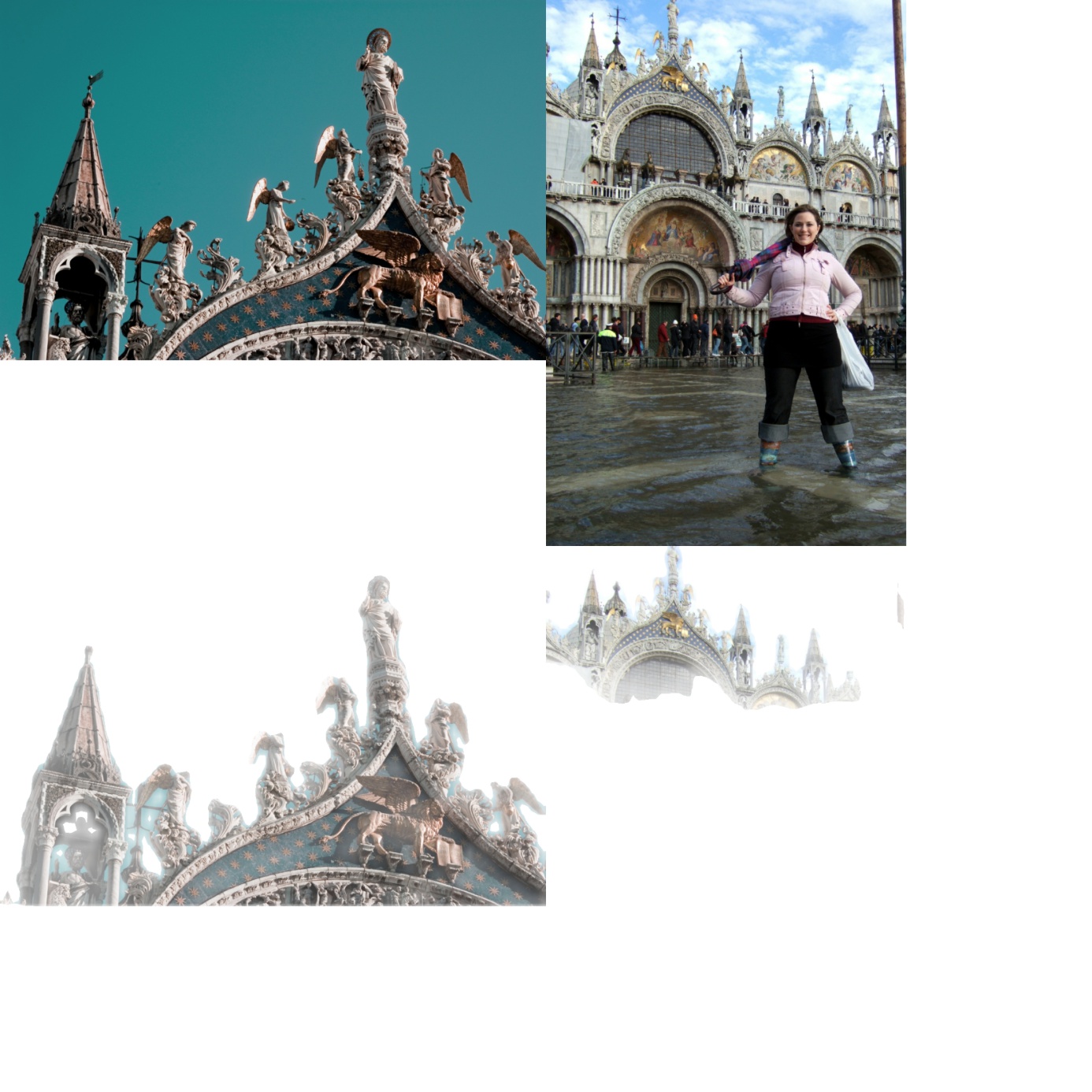}}\\
        \fbox{\includegraphics[width=0.9\linewidth]{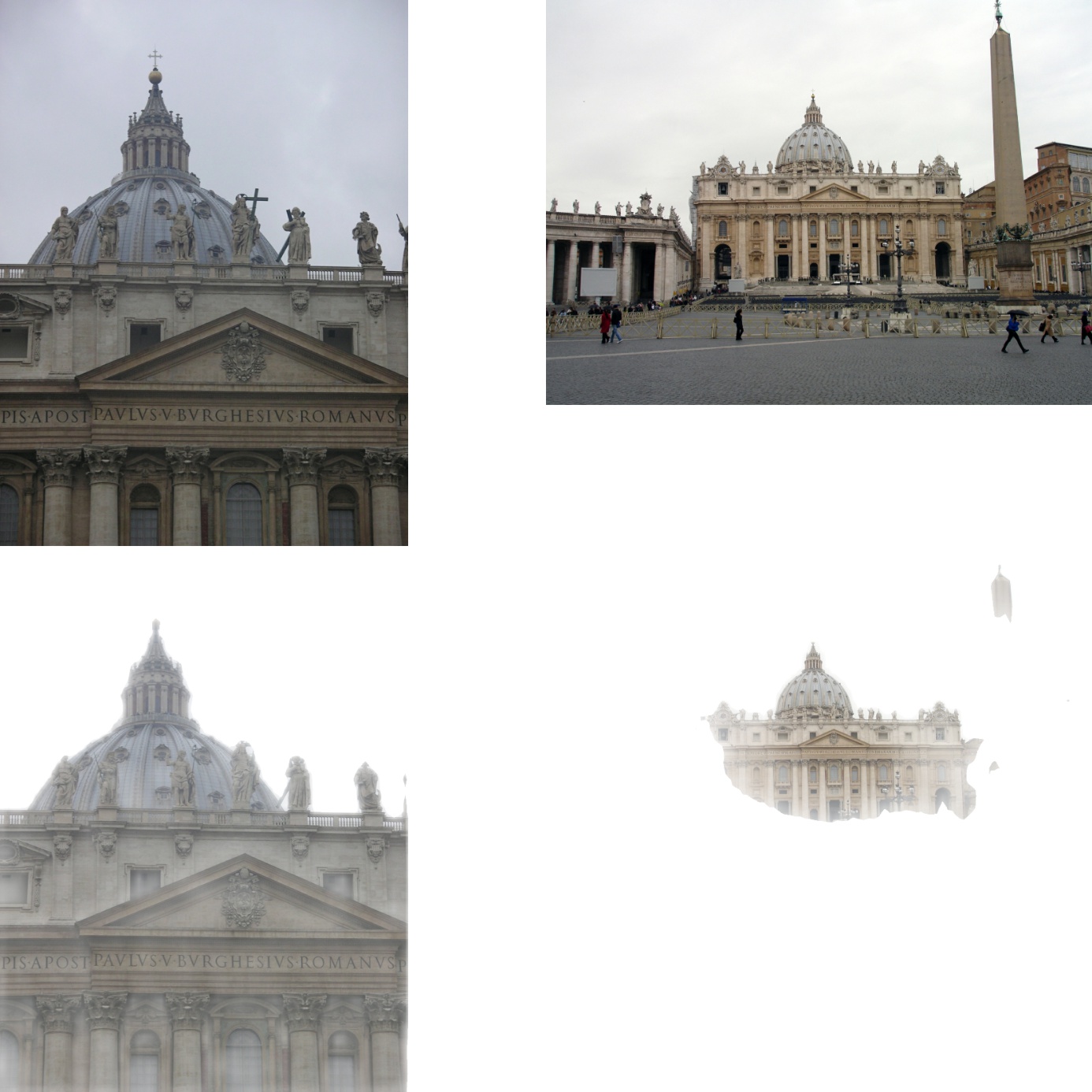}}
 
    \end{minipage}
    \begin{minipage}[b]{0.32\linewidth}
        \fbox{\includegraphics[width=0.9\linewidth]{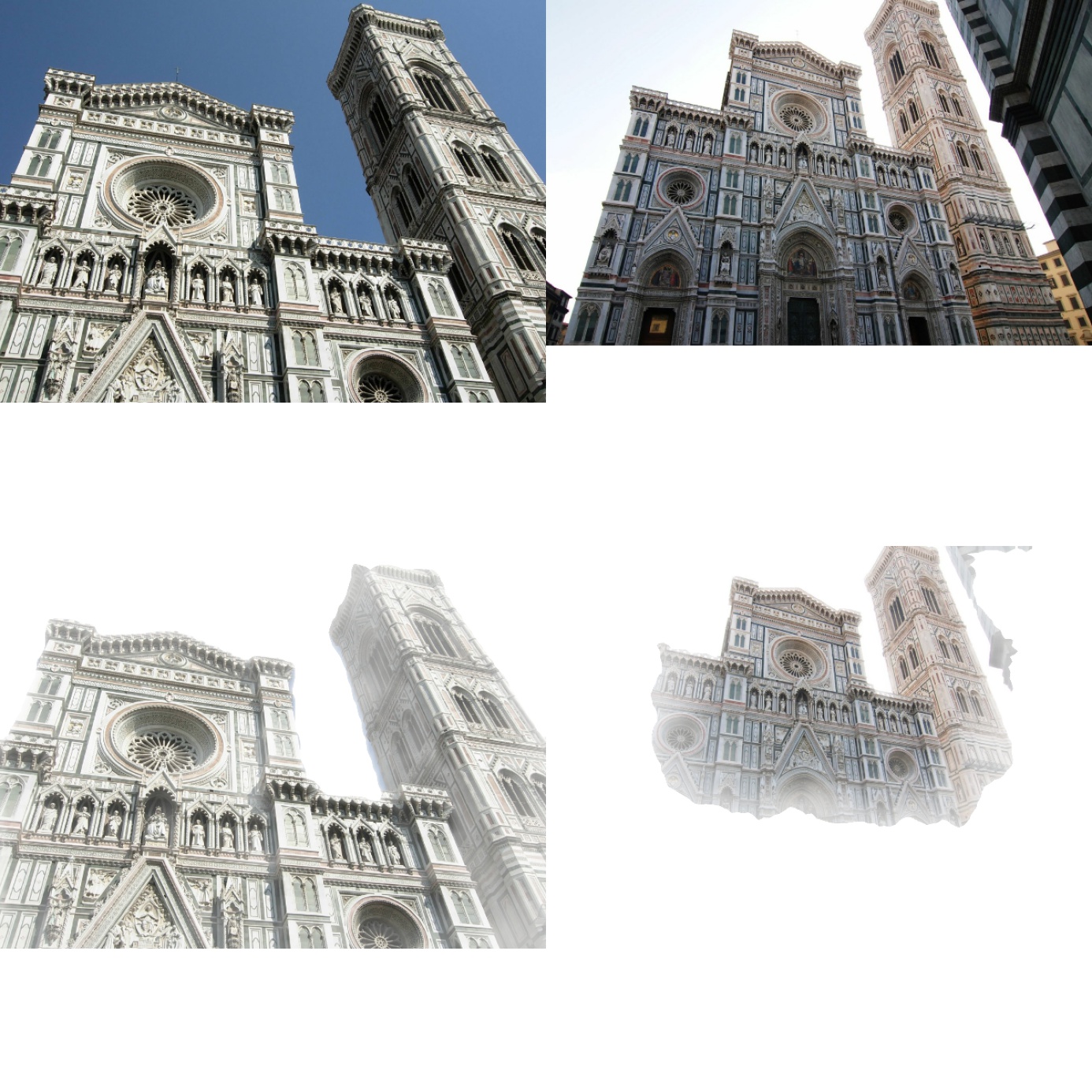}}\\
        \fbox{\includegraphics[width=0.9\linewidth]{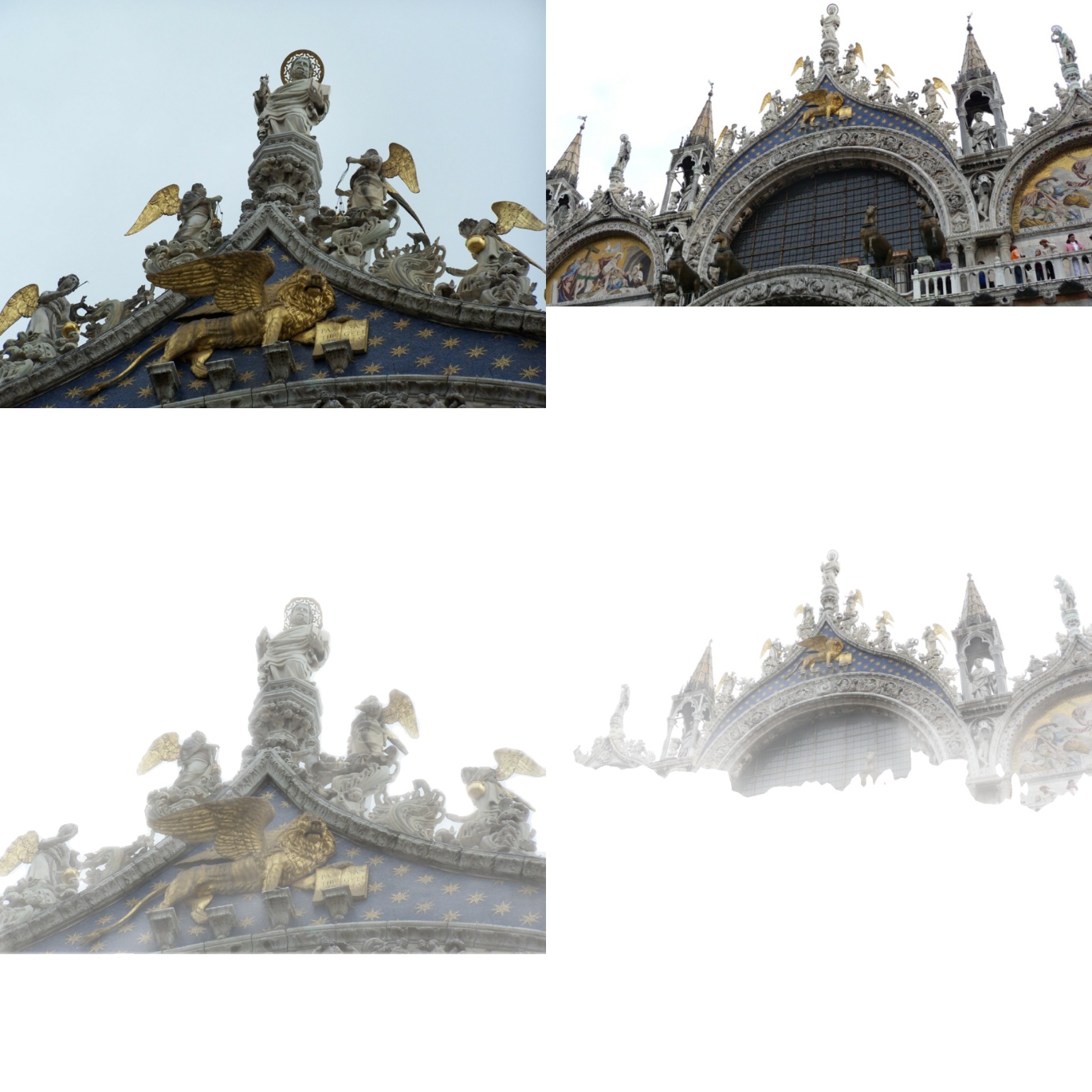}}\\
        \fbox{\includegraphics[width=0.9\linewidth]{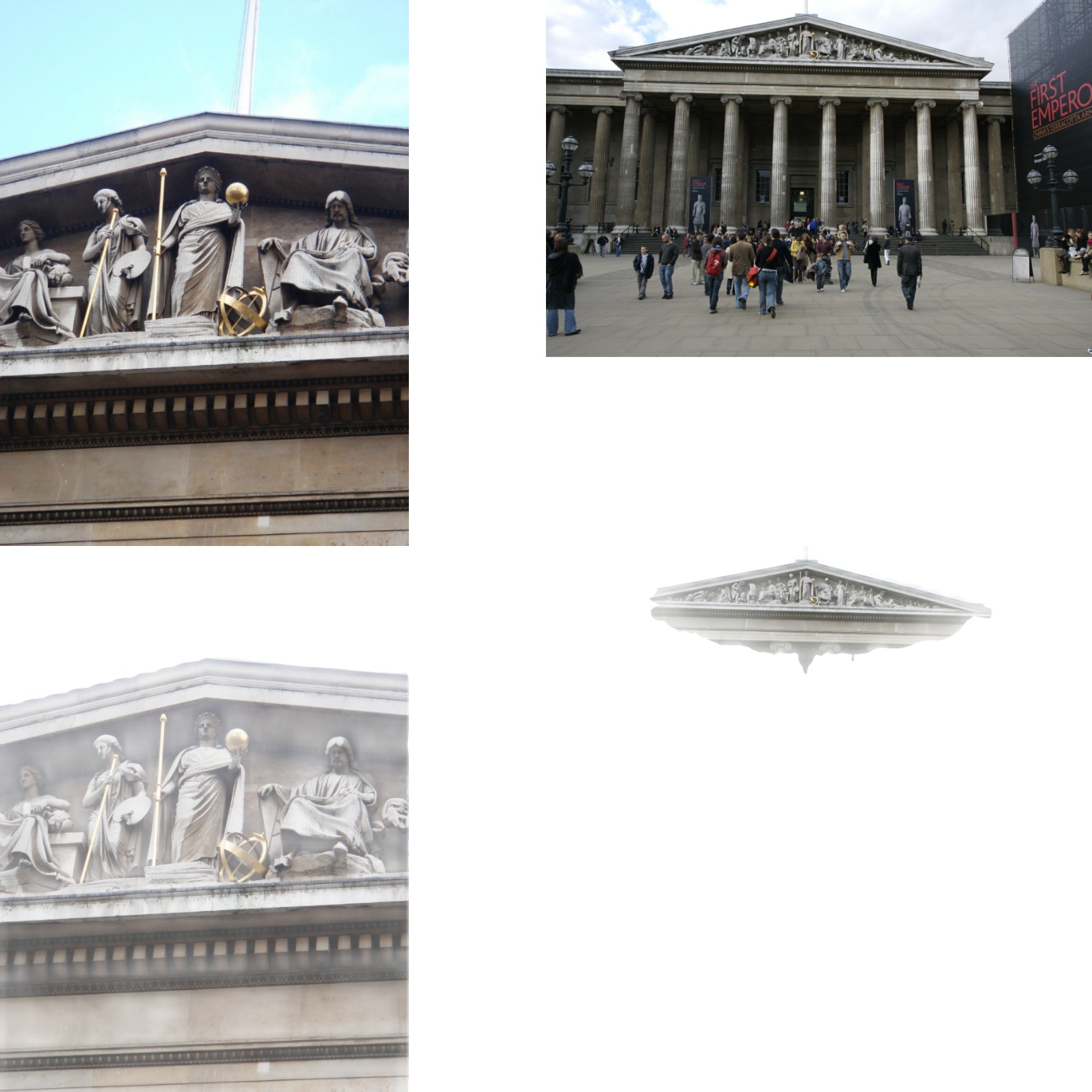}}\\
        \fbox{\includegraphics[width=0.9\linewidth]{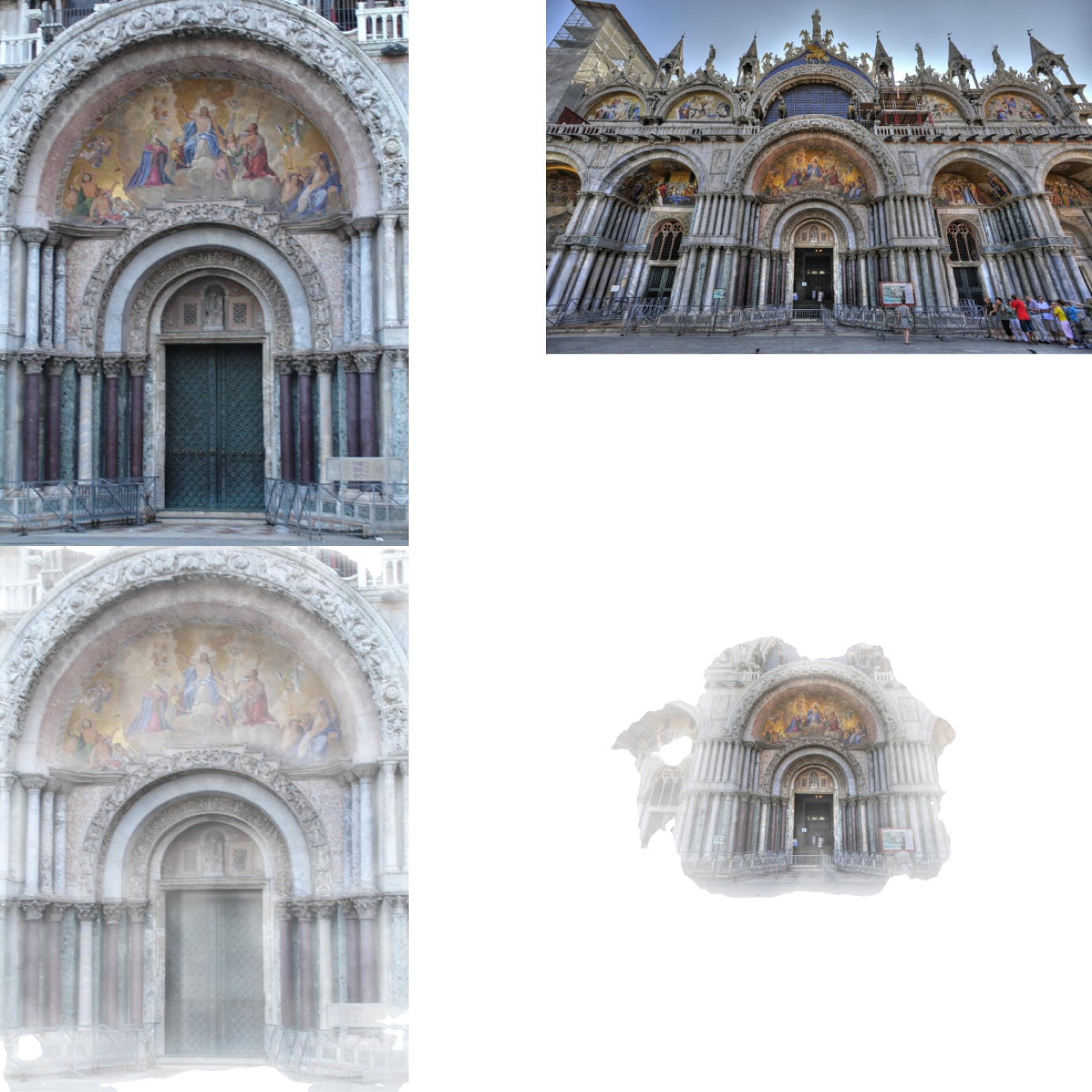}}
    \end{minipage}
  
    \centering
    \caption{Qualitative co-visible area segmentation}
    \label{overlap}
\end{figure*}

We present more qualitative comparisons of AdaMatcher and baselines on Hpatches \cite{Balntas2017} dataset and MegaDepth \cite{Li2018} dataset. In Fig.\ref{hpatch_figure}, we display inlier and outlier matches using different projection thresholds to compare the matching accuracy of different methods on the Hpatches dataset. Fig.\ref{qualitative} presents more qualitative results on the MegaDepth \cite{Li2018} dataset and Fig.\ref{overlap} shows more qualitative results of the co-visible area estimation.

\clearpage
{\small
\bibliographystyle{ieee_fullname}
\bibliography{supplement}
}